\documentclass[lettersize,journal]{IEEEtran}
\usepackage{amsmath,amsfonts}
\usepackage{algorithmic}
\usepackage{algorithm}
\usepackage{array}
\usepackage[caption=false,font=normalsize,labelfont=sf,textfont=sf]{subfig}
\usepackage{textcomp}
\usepackage{stfloats}
\usepackage{url}
\usepackage{verbatim}
\usepackage{graphicx}
\usepackage{cite}
\usepackage{multirow}
\usepackage{booktabs} 
\begin{document}


\title{Triple-CFN: Separating Concepts and Features Enhances Machine Abstract Reasoning Ability}

\author{Ruizhuo Song, Member, IEEE,  Beiming Yuan
\thanks{This work was supported by the National Natural Science Foundation of China under Grants 62273036. Corresponding author: Ruizhuo Song, ruizhuosong@ustb.edu.cn}
\thanks{Ruizhuo Song and Beiming Yuan are with the Beijing Engineering Research Center of Industrial Spectrum Imaging, School of Automation and Electrical Engineering, University of Science and Technology Beijing, Beijing 100083, China (Ruizhuo Song email: ruizhuosong@ustb.edu.cn and Beiming Yuan email: d202310354@xs.ustb.edu.cn). }

\thanks{Ruizhuo Song and Beiming Yuan contributed equally to this work.}
}

\IEEEpubid{0000--0000/00\$00.00~\copyright~2021 IEEE}

\maketitle

\begin{abstract}

This paper proposes a series of innovative frameworks tailored for visual abstract reasoning problems, aiming to enhance the performance of deep learning models in such tasks. This paper highlights the importance of explicitly separating the processes of abstract concept extraction and reasoning feature extraction when designing the architecture of solvers for visual abstract reasoning problems. The validity of this methodology is demonstrated through the superior performance of the Cross-Feature Network (CFN) and its improved version, Triple-CFN. Research indicates that the challenges in visual abstract reasoning not only stem from the more complex induction of abstract patterns compared to traditional visual discrimination tasks but also from conflicts arising from the coexistence of multiple abstract patterns in low-dimensional representations. To more effectively address the such conflicts, the paper introduces a dual Expectation-Maximization (EM) process during the training of the CFN framework. This process alternately optimizes the parameters of the concept extraction module and the feature extraction module within the CFN, shifting its optimization objective to encourage the active synthesis of a set of concepts that are neither conflicting nor detrimental to problem-solving, thereby effectively mitigating conflicts and enhancing the performance of the CFN framework. However, the dual EM process has limitations, potentially leading to overfitting of the synthesized concept set on the training set. To address this, the paper designs mutual information supervision and decorrelation supervision to assist the feature extraction process of the CFN, with experimental results showing the effectiveness of decorrelation supervision. To overcome the limitations of the dual EM process, the paper leverages the metadata accompanying Raven's Progressive Matrices (RPM) problem instances to provide explicit supervisory signals directly to the concept extraction process of the CFN or Triple-CFN, proposing the Meta Triple-CFN. Experimental results reveal that Meta Triple-CFN achieves significant improvements in reasoning accuracy and interpretability on RPM problems. Furthermore, the paper designs a Re-space layer for constructing the feature space during the feature extraction process, and experiments prove that it can further enhance the reasoning accuracy of Triple-CFN, highlighting the importance of constructing a standardized representation space when building reasoning problem solvers. This paper proposes innovative design ideas that provide effective solutions for abstract reasoning problem solvers, with potential benefits across multiple deep learning domains.
Codes are available in: https://github.com/Yuanbeiming/Triple-CFN-Separating-Concepts-and-Features-Enhances-Machine-Abstract-Reasoning-Ability
\end{abstract}


\section{Introduction}

\IEEEPARstart{D}{eep} neural networks have achieved remarkable success in various domains, such as computer vision\cite{ImageNet,ResNet}, natural language processing\cite{Transformer, Bert, GPT-3, Attention in natural language processing, survey of natural language processing}, generative model \cite{GAN,VAE,DiffusionModel}, visual question answering \cite{VQA,CLEVERdataset}, and visual abstract reasoning \cite{RPM,Bongard1,Bongard2}. The advancement of deep learning in this latter field is particularly intriguing and complex. Notably, abstract reasoning aims to equip machines with human-like analytical abilities by learning inherent patterns in data. However, deep learning faces challenges in this area, including complex graphics interpretation, limited reasoning capacity, and generalization difficulties. Overcoming these obstacles demands innovations in model architecture, training techniques, and improved dataset quality.

Visual abstract reasoning problems hold a central position in the field of deep learning. They not only challenge the capabilities of algorithms and Learning system in representation learning and feature extraction but also drive the development of model generalization, interpretability, and transparency. These problems demand that deep learning models and Learning system capture abstract concepts and relationships within images and perform logical reasoning and decision-making based on this foundation.

Moreover, visual abstract reasoning problems have advanced the study of generalization capabilities in deep learning models, requiring them to learn from limited data and adapt to new situations. This necessitates not only attention to the surface features of the data but also an understanding of the underlying logical structures. Additionally, to enhance model interpretability, researchers have explored methods such as attention mechanisms and neuro-symbolic reasoning, attempting to simulate human reasoning processes and provide insights into model decision-making.
These issues have also stimulated innovation in deep learning algorithms, promoting the construction of intelligent systems that more closely resemble human cognition. 
Large models, as one of the most popular research directions currently, have encountered setbacks in their exploration within this field, which has attracted widespread attention in the academic community \cite{LLM1, LLM2, LLM3}. This phenomenon undoubtedly underscores the profound significance and important value inherent in this research area.

In summary, visual abstract reasoning problems play a crucial role in advancing deep learning towards higher levels of intelligence and cognitive capabilities. We firmly believe that the technological and theoretical innovations stemming from these reasoning tasks have the potential to bring significant benefits to various AI domains, including large models.
Currently, notable abstract reasoning problems worthy of investigation include: Ravens Progressive Matrices (RPM) problems\cite{RPM} and Bongard problems\cite{Bongard1,Bongard2}. They present learning demands that span from perception to reasoning. Addressing these problems demands necessitates advancements in deep learning's ability to effectively handle visual abstract reasoning tasks.

\subsection{RAVEN Database}

The RAVEN database\cite{RAVENdataset} presents a unique challenge in the realm of RPM problems, with each question typically comprising 16 images enriched with geometric entities. Half of these images, specifically 8, form the problem stem while the remaining 8 constitute the answer pool. Subjects are tasked with selecting appropriate images from the answer pool to complete a 3×3 matrix, following aprogressive pattern of geometric images along the rows to convey specific abstract concepts.

\begin{figure}[htp]\centering
	\includegraphics[trim=10cm 0cm 0cm 0cm, clip, width=7cm]{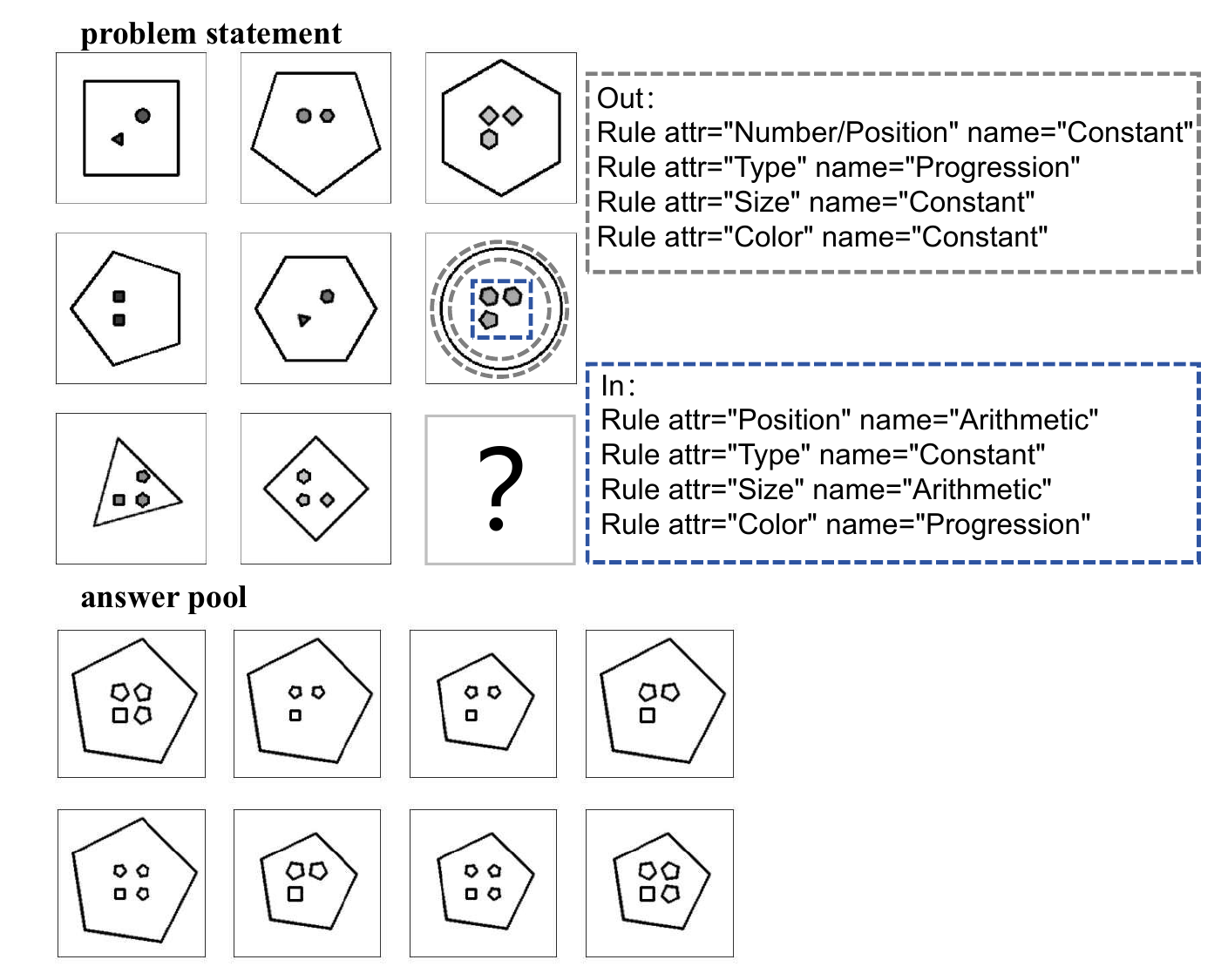}
	\caption{RAVEN case}
\label{RAVEN case}
\end{figure}

As illustrated in figure \ref{RAVEN case}, the construction of a RAVEN problem speaks to its generality and sophistication. Within these problems, certain human-defined concepts within the geometric images, such as ``shape" or ``color", are deliberately abstracted into bounded, countable, and precise ``visual attributes". The notion of ``rule" is then employed to delineate the progressive transformation of a finite set of these visual attribute values. However, it's worth noting that some visual attributes remain freedom of the rule, potentially posing as distractions for deep model reasoning.

The RVEN database is further diversified into multiple sub-databases, namely: single-rule groups—center single (center), distribute four (G2×2), distribute nine (G3×3)—and dual-rule groups: in center single out center single (O-IC), up center single down center single (U-D), left center single up center single (I-L), and in distribute four out center single (O-IG). In problems with a singular rule, the progressive transformation of an entity's attributes within the image adheres to one set of rules, while in those with dual rules, two independent rule sets govern this transformation.

\subsection{PGM Database}
\begin{figure}[htp]\centering
	\includegraphics[trim=8cm 0cm 0cm 0cm, clip, width=7cm]{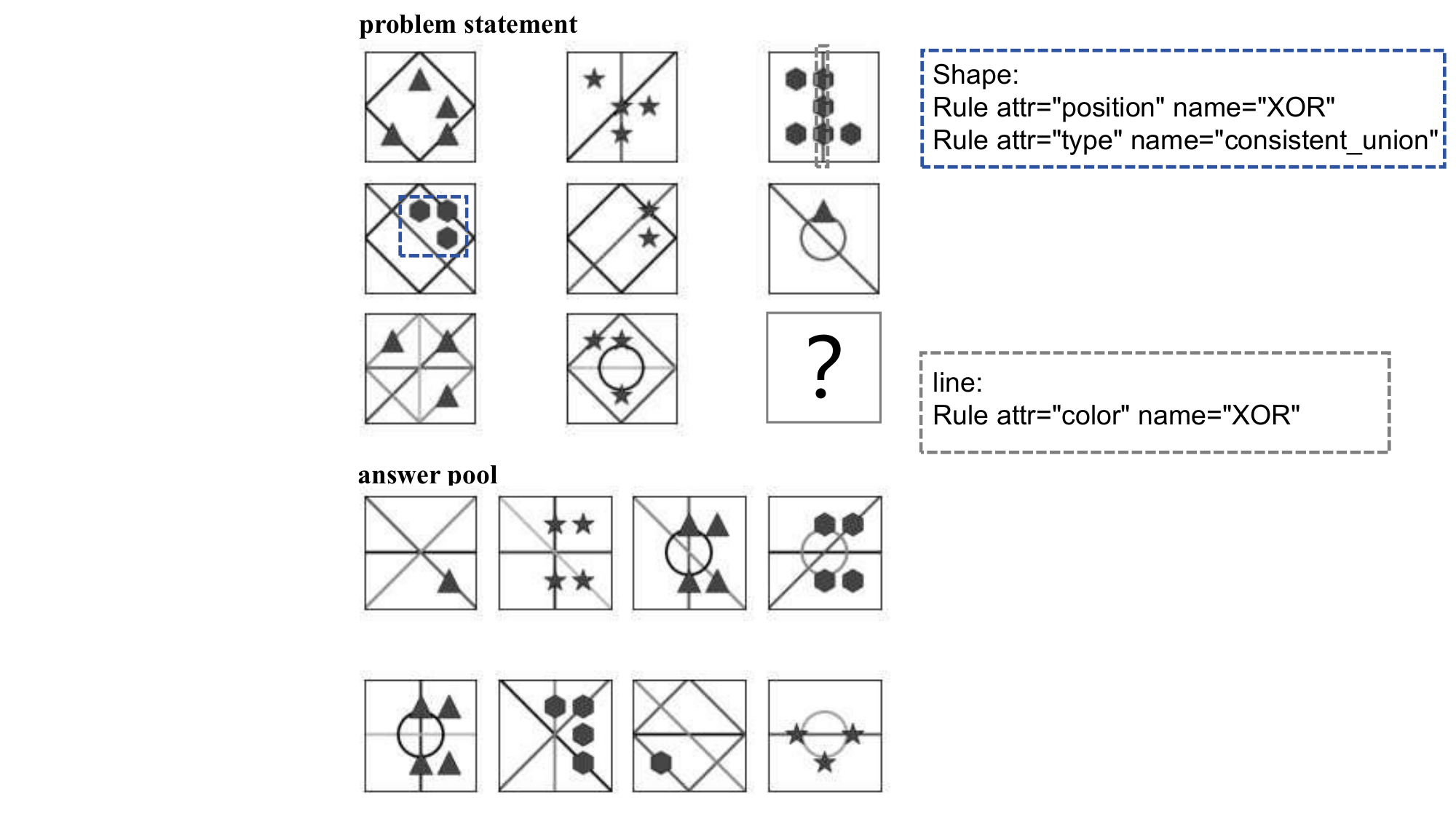}
	\caption{PGM case}
\label{PGM case}
\end{figure}

PGM\cite{PGMdataset} problem is also a kind of RPM problems. Its extremely high difficulty has earned him a renowned reputation.
The design logic of PGM and RAVEN problems is similar, with both types of problems represented by a problem stem composed of 8 images and an answer pool formed by another 8 images. An example of a PGM problem is illustrated in the provided figure \ref{PGM case}.

Consequently, the difficulty of RPM problems lies not only in the exploration of visual attributes at different levels but also in the induction and learning of the progressive patterns of ``visual attributes."

\subsection{Bongard-logo Database}
\begin{figure}[htp]\centering
	\includegraphics[trim=7cm 0cm 7cm 0cm, clip, width=6cm]{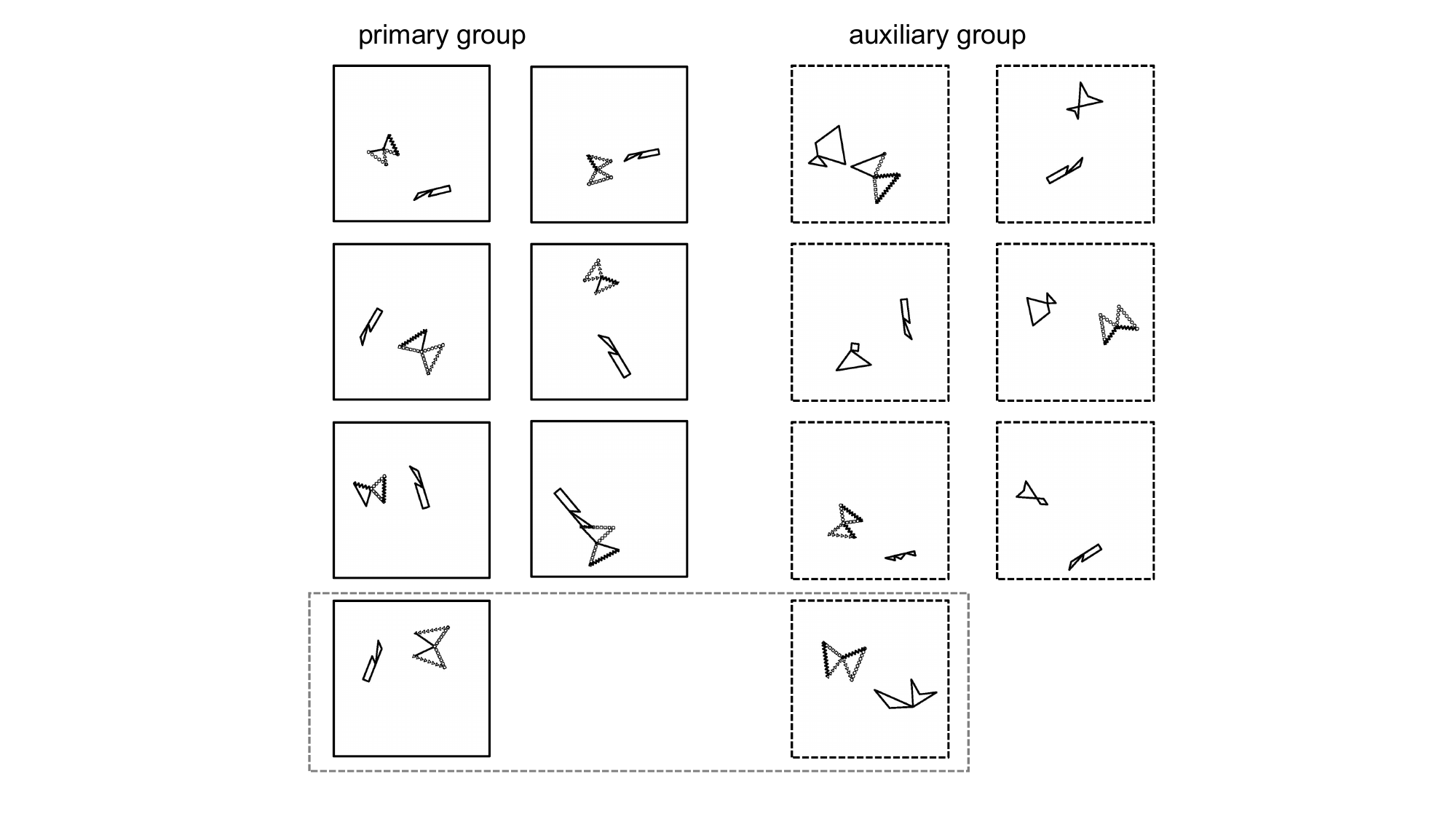}
	\caption{Bongard case}
\label{Bongard case}
\end{figure}

Bongard problems\cite{Bongard1} exhibit significant differences from RPM problems, with Bongard problems being a type of small sample learning problem and clustering reasoning problem\cite{clustering}. Typically composed of multiple images, these problems divide the images into two groups: a primary group and a auxiliary group. All images within the primary group express abstract concepts constrained by certain rules, while the images in the auxiliary group reject these rules to varying degrees. Bongard problems challenge deep learning algorithms to correctly categorize ungrouped images into the appropriate small groups. Bongard-logo\cite{Bongard2}, an instantiation of Bongard problems within the realm of abstract reasoning, poses considerable reasoning difficulties. Each Bongard-logo problem consists of 14 images, with 6 images in the primary group, 6 in the auxiliary group, and the remaining 2 serving as options for grouping. The images contain numerous geometric shapes, and their arrangements serve as the basis for grouping. Figure \ref{Bongard case} illustrates an example Bongard-logo problem. In Figure \ref{Bongard case}, each Bongard problem is composed of two sets of images: the primary group A and the auxiliary group B. The primary group A contains 6 images, with the geometric entities within each image following a specific set of rules, while the auxiliary group B includes 6 images that reject the rules in group A. The task is to determine whether the images in the test set satisfy the rules expressed by group A. The difficulty level varies depending on the problem's structure.

The Bongard-logo problem suite evaluates model intelligence across four dimensions:
\begin{enumerate}
    \item \textbf{The Combinatorial Abstract Shape Test Set (CM)}  
    Assesses logical reasoning by requiring systems to identify novel attribute combinations never seen during training, testing their ability to derive abstract patterns from known elements.
    
    \item \textbf{The Novel Abstract Shape Test Set (NV)}  
    Employs attribute withholding methodology to evaluate innovative potential, examining models' ability to infer entirely new rules when one attribute and all its combinations are systematically excluded from training.
    
    \item \textbf{The Basic Shape Test Set (BA)}  
    Measures adaptability to novel configurations of fundamental geometric primitives, focusing on parameter variations like quantity, position, and orientation.
    
    \item \textbf{The Free-form Shape Test Set (FF)}  
    Presents the greatest challenge through unstructured natural forms, evaluating comprehensive visual processing capabilities in recognizing organic patterns beyond rigid templates.
\end{enumerate}
This evaluation framework requires models to first train on standardized datasets before progressing through these hierarchically structured tests, enabling comprehensive assessment from basic pattern recognition to advanced abstract reasoning.

\section{Related work}

\subsection{RPM solver}

In the realm of abstract visual reasoning, innovative approaches have been developed to transcend the constraints of conventional models. For example, CoPINet\cite{CoPINet} refines attribute determination through contrastive learning of graphic nuances. LEN+teacher\cite{LEN} employs a student-teacher framework to enhance knowledge transfer and model optimization. DCNet\cite{DCNet} boosts accuracy and efficiency with a dual-contrast module that integrates multiple reasoning factors. NCD\cite{NCD} improves generalization with pseudo-targets and decentralized strategies, while SCL\cite{SCL} addresses complex tasks through multiple monitoring mechanisms. SAVIR-T\cite{SAVIR-T} enhances reasoning efficiency by integrating multi-perspective information, and RS-Tran\cite{RS} refines RPM predictions with multi-view, multi-evaluation techniques. CRAB\cite{CRAB} excels in specific databases, and studies show that decoupled perceptual features\cite{RAVEN solver} and symbolic methods\cite{PrAE,ALANS,NVSA} significantly bolster reasoning capabilities. These advancements offer robust tools for visual reasoning research and applications.

\subsection{Bongard-logo solver}

In tackling Bongard problems, the academic community has examined several strategies. The language-based feature model \cite{Bongard1} interprets visual features through a formal language but struggles with abstract concepts and requires retooling for new problems. The convolutional neural network model \cite{Bongard3}, while popular, is heavily reliant on the quality and volume of training data. Synthetic data techniques \cite{Bongard2} aim to improve generalization but are constrained by the relevance of the data. Notably, the PMoC \cite{PMoC} model offers an effective solution for Bongard-Logo problems with its standalone probability model. Overall, each method has its merits and demerits, and future research should aim for more holistic and integrated approaches to Bongard problems.



\subsection{Covariance matrix and correlation loss}

The covariance matrix stands as a pivotal tool in multivariate statistical analysis, quantifying the relationships between multiple random variables\cite{ViCReg}. Treating image representation dimensions as distributions, the covariance matrix assesses linear correlations within image data, aiding downstream image analysis and processing tasks. Formulas (\ref{matrix_cov}) and (\ref{eq_cov}) are used to calculate the covariance matrix and correlation loss, respectively.

\begin{align}
M_\sigma(x) =& \frac{1}{N-1} \sum_{i=1}^{N} (\mathbf{x}_i - \bar{\mathbf{x}})(\mathbf{x}_i - \bar{\mathbf{x}})^\top\label{matrix_cov}\\
\ell(x)=&\frac{1}{d} \sum \left(M_\sigma(x)^2\cdot (1-I)\right)\label{eq_cov}
\end{align}
Where $I$ denotes the identity matrix, and $M_\sigma(x) \in R^{d \times d}$. $d$ represents the dimensions of the vector $x_i$, and $n$ refers to the number of samples involved in the computation, given that the covariance matrix is calculated based on a batch of samples.

\subsection{The Expectation-Maximization (EM) algorithm}
The Expectation-Maximization (EM) algorithm\cite{EM} represents a powerful iterative method widely employed in statistics for finding maximum likelihood estimates of parameters in probabilistic models, especially when the data contain missing values or are observed in an incomplete manner. By alternating between an expectation ``E" step and a maximization ``M" step, the algorithm optimizes the likelihood function, gradually refining parameter estimates until convergence. 





\section{Methodology}

Visual abstract reasoning problems, such as RPM\cite{RPM} and Bongard problems\cite{Bongard1,Bongard2}, constitute advanced discriminative visual tasks that demand the cognition and identification of abstract patterns in visual information, exceeding the requirements of typical discriminative tasks. Progress in these visual reasoning tasks often guides conventional discriminative tasks, prompting increased research in this area in recent years.

Most studies are enthusiastic about designing deep neural networks to encode the features and attributes of reasoning images into abstract representations, and thereby solve reasoning tasks, which is also a common strategy in discriminative visual tasks. However, this paper argues that the methodology, which prevails in typical discriminative tasks, is inadequate. The setbacks encountered in the current research on large models for visual abstract reasoning problems also indicate the shortcomings of this methodology \cite{LLM1, LLM2, LLM3}. The key to solving complex visual reasoning tasks lies in separately extracting features and concepts, and leveraging their interactions for reasoning. This paper revolves around this core proposition and methodology.

\section{A baseline for Bongard-Logo}\label{section baseline}

\begin{figure}[htp]\centering
	\includegraphics[trim=2cm 0cm 13.5cm 0cm, clip, width=6.5cm]{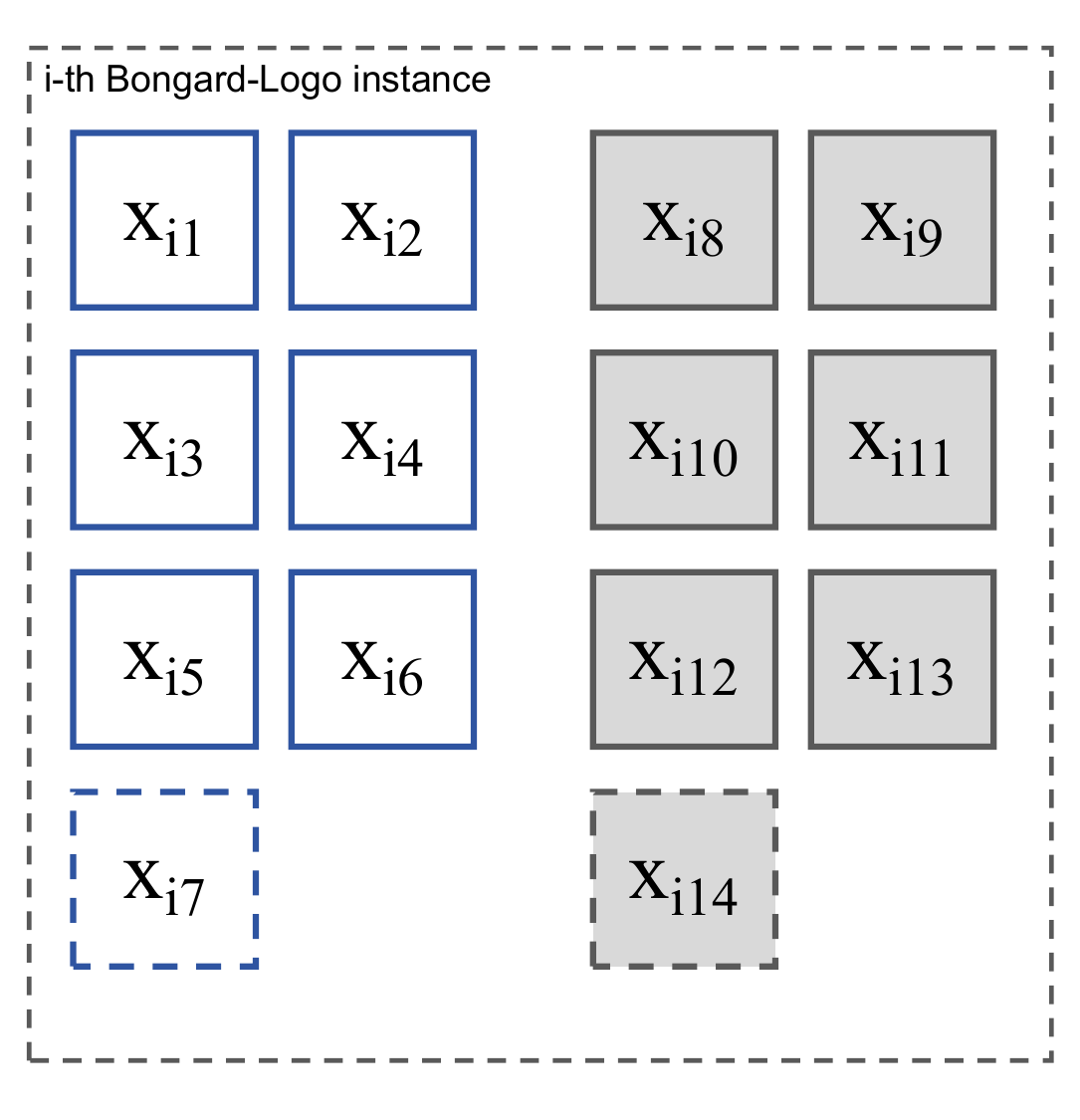}
	\caption{Feedforward process baseline}
\label{Bongard-Logo_label}
\end{figure}

Based on higher-dimensional human concepts and preferences, the creators of Bongard-logo problems have categorized the Bongard-logo dataset into four distinct problem types: FF, BA, NV, CM. 
For the purpose of conveniently marking images in the Bongard-Logo instance, we denote the images in Bongard-Logo instances as $x_{ij}$ according to Figure \ref{Bongard-Logo_label}, where $i$ represents the problem's identifier, with $i\in[1,n]$ and $n$ signifying the total number of instances. Specifically, $\{x_{ij}|\,j\in[1,6]\}$ represents images in the $i$-th primary group (positive instances), while  $\{x_{ij}|\,j\in[8,13]\}$ represents images in the $i$-th auxiliary group (negative instances). Additionally, $x_{i7}$ represents the test image to be potentially assigned to the $i$-th primary group, and $x_{i14}$ represents the test image to be potentially assigned to the $i$-th auxiliary group.

\subsection{Structure of the Baseline}
\begin{figure}[htp]\centering
	\includegraphics[trim=2cm 0cm 14cm 0.5cm, clip, width=8.5cm]{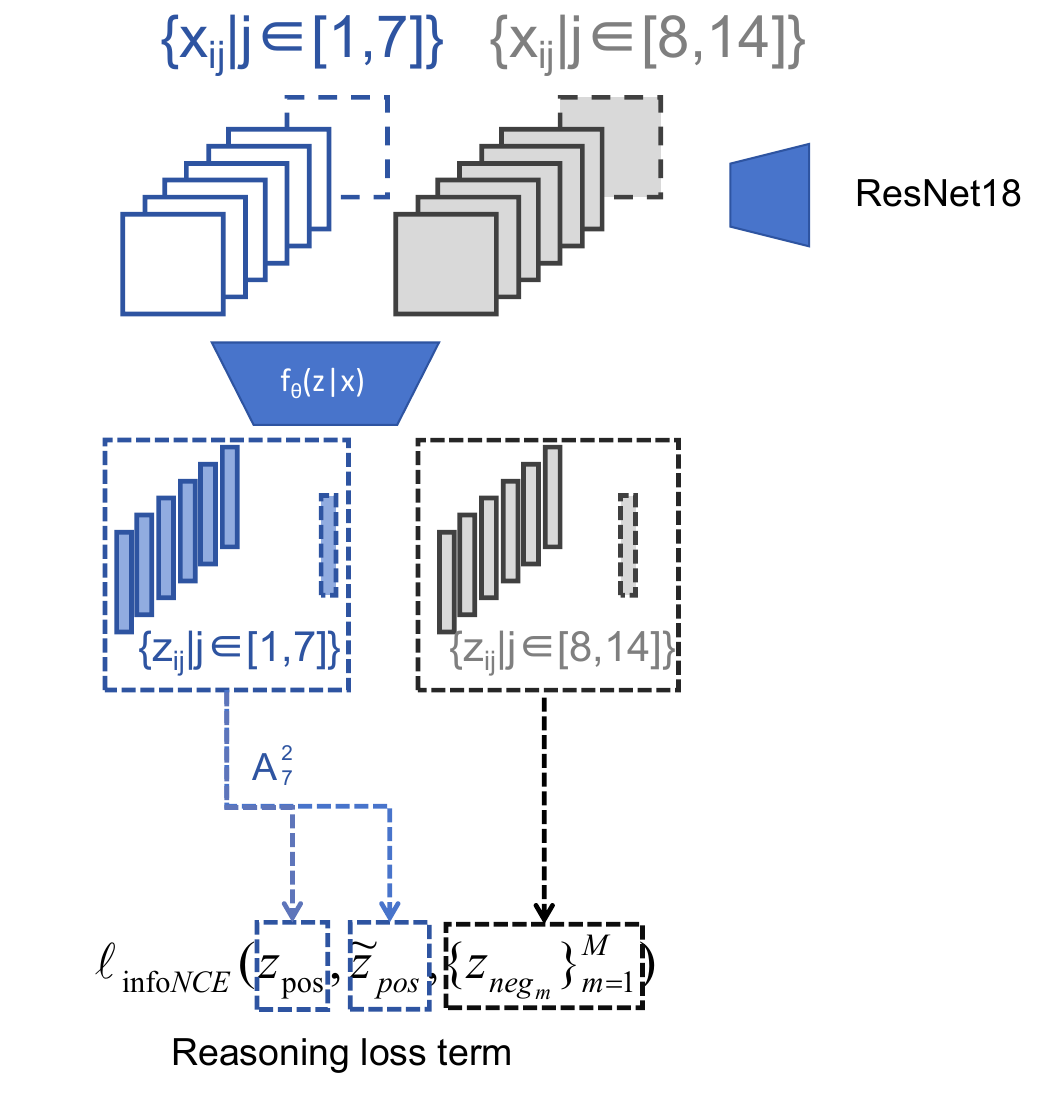}
	\caption{Feedforward process baseline}
\label{baseline}
\end{figure}

In this paper, we utilize the InfoNCE loss function \cite{InfoNCE} as a reasoning loss term, with the aim of training a standard ResNet18 network to solve the Bongard-Logo problem. The resulting ResNet18 model, denoted as $f_\theta (z|\,x)$, is capable of mapping an image $x_{ij}$ to its latent representation $z_{ij}$, and possesses the ability to handle individual or concurrent high-dimensional concepts inherent in the Bongard-logo dataset. The reasoning loss term based on InfoNCE can be formalized as follows:
\begin{align}
    &{\ell _\mathbf{InfoNCE}}({z_{pos}},{\tilde z_{pos}},\{ {z_{ne{g_m}}}\} _{m = 1}^M) \nonumber\\
    &=  -\log \frac{{{e^{({z_{pos}} \cdot {\tilde z_{pos}})/t}}}}{{{e^{({z_{pos}} \cdot {\tilde z_{pos}})/t}} + \sum\nolimits_{m = 1}^M {{e^{({z_{pos}} \cdot {z_{ne{g_m}}})/t}}} }}\label{InfoNCE}
\end{align}
Specifically, $z_{pos}, \tilde{z}_{pos} \in \{z_{ij}, \overline{z}_i \mid j \in [1,7]\}$, where $\overline{z}_i = \frac{1}{7} \sum_{j=1}^7 z_{ij}$ and $z_{pos} \neq \tilde{z}_{pos}$. Additionally, $\{z_{\text{neg}_m}\}_{m=1}^M = \{z_{ij} \mid j \in [8,14]\}$ and $t = 10^{-3}$. $\overline{z}_i$ is used for contextual analysis within the primary group. InfoNCE can constrain the encoding capability of $f_\theta(z | x)$ such that it encodes the Bongard-Logo image set $\{x_{ij} | j \in [1,14]\}$ into the representation set $\{z_{ij} | j \in [1,14]\}$, and the cosine similarity within the set $\{z_{ij} | j \in [1,7]\}$ is greater than the cosine similarity between the set $\{z_{ij} | j \in [1,7]\}$ and the set $\{z_{ij} | j \in [8,14]\}$. Such $f_\theta(z | x)$ conforms to the problem-solving logic of the Bongard-Logo problem. We can solve the Bongard-Logo problem by measuring the cosine similarity between representations.
The aforementioned training process can be represented as Figure \ref{baseline}. In Figure \ref{baseline}, ``$A_7^2$" represents the number of ways to choose 2 elements from a set of 7 elements.

\subsection{Preliminary Experiments on the Bongard-Logo Problem}

Utilizing the aforementioned methodologies, this study conducted experiments on the four concept databases of Bongard-Logo, both individually and in combination. The experimental results are presented in Table \ref{DC_Bongard_Results}. In this table, the ``Bongard-Logo" entry encompasses the findings obtained by training the model using a combined dataset of all four concepts. Conversely, the ``Separated Bongard-Logo" entry documents the outcomes derived from training each concept independently.

\begin{table}[h]
\caption{Reasoning Accuracies of ResNet18 on Bongard-logo.}
\label{DC_Bongard_Results}
\centering
\begin{tabular}{cccccc}
\toprule
&\multicolumn{4}{c}{Test Accuracy(\%)}& \\
Data Set& FF&BA&CM&NV \\
\midrule
 Bongard-logo&88.1&97.9&76.0&75.8\\
\midrule
Separated Bongard-logo&97.9&99.0&75.0&72.8\\
\bottomrule
\end{tabular}
\end{table}

The results clearly indicate that the convolutional deep model $f_\theta (z|\,x)$ faces challenges in distinguishing between four different concepts. 

\subsection{High-dimensional human concepts conflict with each other in low-dimensional representations.
}
Convolutional Neural Networks (CNNs) are capable of abstracting pixel configurations into representations tailored to the task \cite{A survey of convolutional neural networks}. However, in the Bongard-Logo problem, multiple tasks or concepts intertwine, complicating the scenario. Different concepts impose diverse task requirements on the network model $f_\theta (z|\,x)$, compelling it to store multiple distinct problem-solving methods. However, current experimental observations indicate that these different concepts are forcing $f_\theta (z|\,x)$ to maintain conflicting image abstraction modes. In other words, these four concepts exhibit conflicts when generating abstract representations, leading to confusion within the neural network and making it difficult to accurately distinguish among the four concepts.

From a different perspective, the image representations encoded by CNNs, which are competent for the Bongard-Logo problem, can be interpreted as belonging to a conditional distribution subject to two factors: pixel configuration and concept. This distribution can be expressed as ``$P(\text{representation}|\,\text{pixel configurations},\text{concept})$". Clearly, in the absence of concept hints, CNNs that rely solely on image pixel configurations as input are unable to effectively model such a conditional distribution.

Therefore, this paper argues that a competent solver for the Bongard-Logo problem should be able to infer concepts from image pixel configurations and solve problems based on the inferred concepts.

\section{A Framework for the RPM and Bongard Problems
}
In this section, inspired by the characteristics of the Bongard-Logo problem, this paper designs the CFN network, followed by the dual EM algorithm and the Triple-CFN to address the shortcomings present in the CFN.

\subsection{Cross-Feature Net (CFN)}

Based on the analysis of the characteristics of the Bongard-Logo visual abstract reasoning problem presented in the previous section, this paper argues that it is necessary to separately extract concepts and features from the reasoning problem. Subsequently, by computing the interactions between these concepts and features, a more reasonable final representation capable of undertaking the reasoning task can be obtained.

\begin{figure}[htp]\centering
	\includegraphics[trim=3cm 3cm 2cm 2cm, clip, width=8cm]{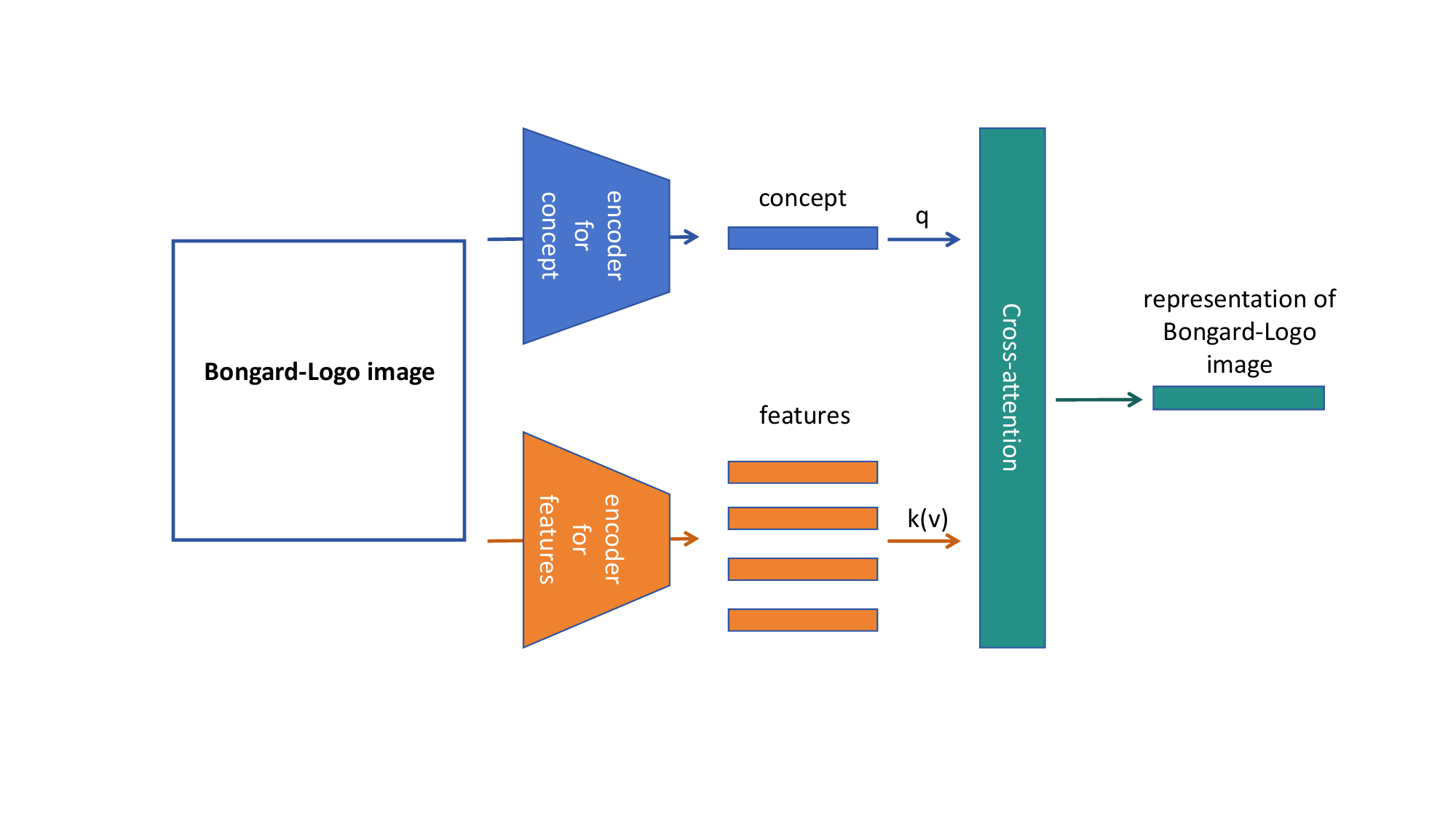}
	\caption{The framework of the Bongard-Logo solver where the concept and feature extraction processes are separated.}
\label{Framework}
\end{figure}

Specifically, we use two networks to separately extract concepts and features from instances of the reasoning problem, with both concepts and features essentially being latent variables of the instances. Subsequently, we use the encoded concepts as queries and the encoded features as key-value pairs. By calculating the cross-attention results between them, we obtain a representation that reflects the interaction between the concepts and the features. 
The effectiveness of the attention mechanism in analyzing complex logic is widely recognized. \cite{Attention in natural language processing}
This framework can be expressed as Figure \ref{Framework}.
After establishing the framework, there are still some details that need to be further determined and clarified.

In detail, our framework comprises three modules: $g_\omega(q_{ij}|x_{ij})$, $g_\theta(k_{ij}|x_{ij})$, and $g_\eta(z_{ij}|q_{ij},k_{ij})$. The module $g_\omega(q_{ij}|x_{ij})$ is responsible for encoding the Bongard-Logo image $x_{ij}$ into a concept vector $q_{ij}$. Similarly, $g_\theta(k_{ij}|x_{ij})$ encodes $x_{ij}$ into a feature vector $k_{ij}$. Lastly, $g_\eta(z_{ij}|q_{ij},k_{ij})$ calculates the cross-attention between $q_{ij}$ and $k_{ij}$, thereby obtaining the final representation $z_{ij}$ of the image $x_{ij}$. $g_\omega(q_{ij}|x_{ij})$, $g_\theta(k_{ij}|x_{ij})$, and $g_\eta(z_{ij}|q_{ij},k_{ij})$ correspond to the blue, orange, and green modules in the framework shown in Figure \ref{Framework}, respectively.

\begin{figure}[htp]\centering
	\includegraphics[trim=2cm 0cm 7cm 0cm, clip, width=8.0cm]{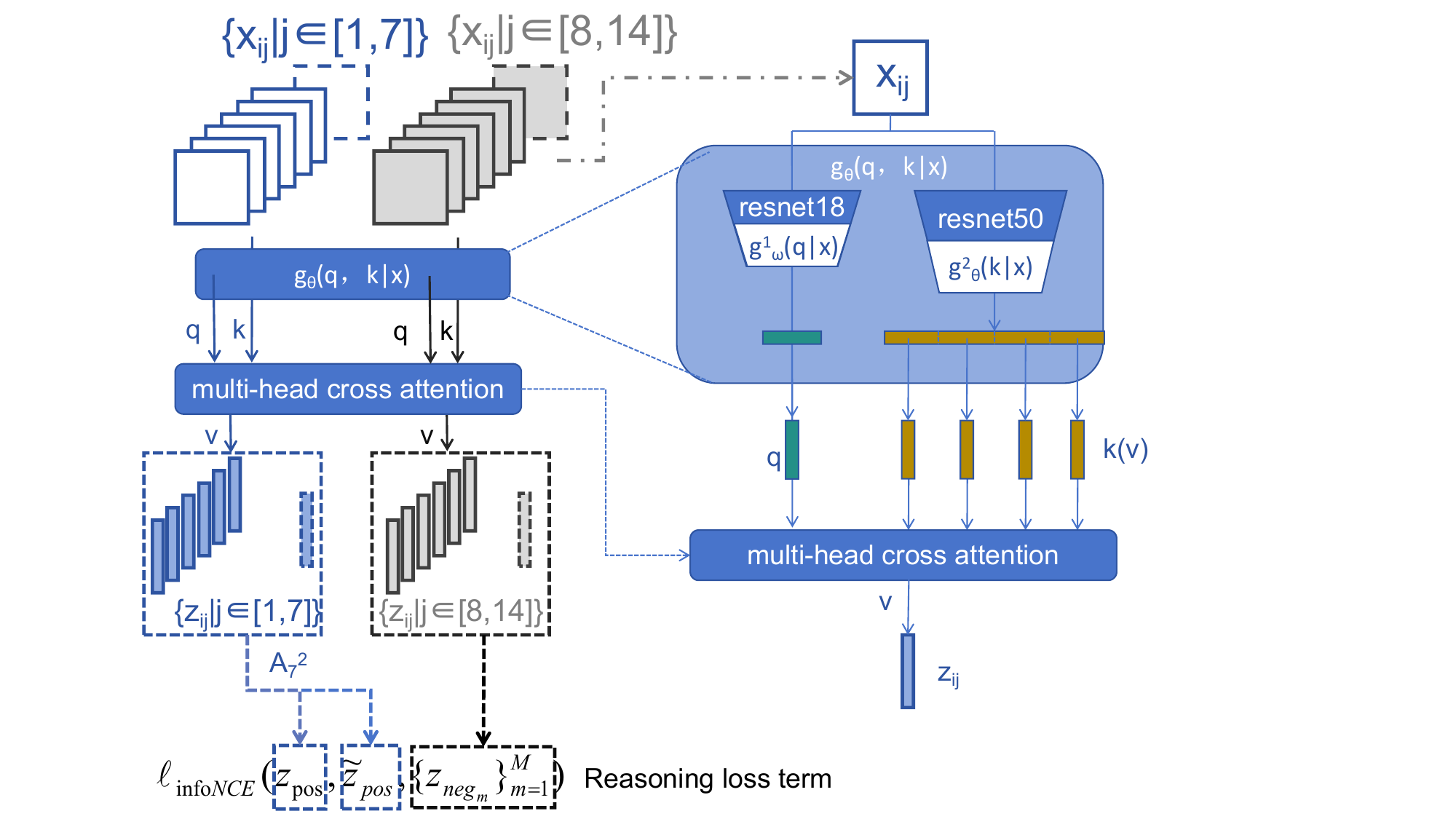}
	\caption{structure of CFN}
\label{CFN}
\end{figure}

We employ ResNet18 and ResNet50 as the architectures for $g_\omega(q_{ij}|x_{ij})$ and $g_\theta(k_{ij}|x_{ij})$, respectively, while the Transformer-Decoder serves as $g_\eta(z_{ij}|q_{ij},k_{ij})$. We train this three-module network using the InfoNCE loss, which we refer to as the Cross-Feature Net (CFN). The feedforward process of the CFN is shown in Figure \ref{CFN}. This paper concerns that the pattern of CFN, which decodes the final representation from features using concepts, is the key for deep learning algorithms to solve abstract reasoning problems.
However, the current structure of the CFN is still alluring on the outside but flawed within.

\subsection{Dual Expectation-Maximization Process}

Experiments in Section \ref{section baseline} indicate that concepts in Bongard-Logo conflict at the representation level, questioning the validity of the methodology of extracting concepts separately and challenging the effectiveness of the CFN structure. Therefore, we hypothesize the existence of a concept set \( \{Y_\alpha\} \), where each \( Y_\alpha \) represents a unique concept that may differ from the original framework of Bongard-Logo. These concepts are assumed to be non-conflicting and do not alter the original solutions of Bongard-Logo.
When such a set of concepts exists, our CFN is deemed rational and efficacious.
However, it is insufficient to merely postulate the existence of a set of concepts $\{Y_\alpha\}$ without actively seeking and fitting such a set of concepts.

It is impractical and meaningless for us to artificially design such a set of concepts $\{Y_\alpha\}$, as it is too manual and lacks intelligence. Therefore, we attempt to encourage the CFN to independently deduce such a set of concepts $\{Y_\alpha\}$.
This paper argues that the task of prompting the CFN to search for a set of concepts that can decode given image features into optimal final representations that maximize the solution of Bongard-Logo problems bears a resemblance to the Expectation-Maximization (EM) process. Consequently, this paper contends that this task can be effectively accomplished by integrating the EM process into the training regimen of the CFN.

\begin{figure}[htp]\centering
	\includegraphics[trim=3cm 1cm 1cm 1cm, clip, width=8cm]{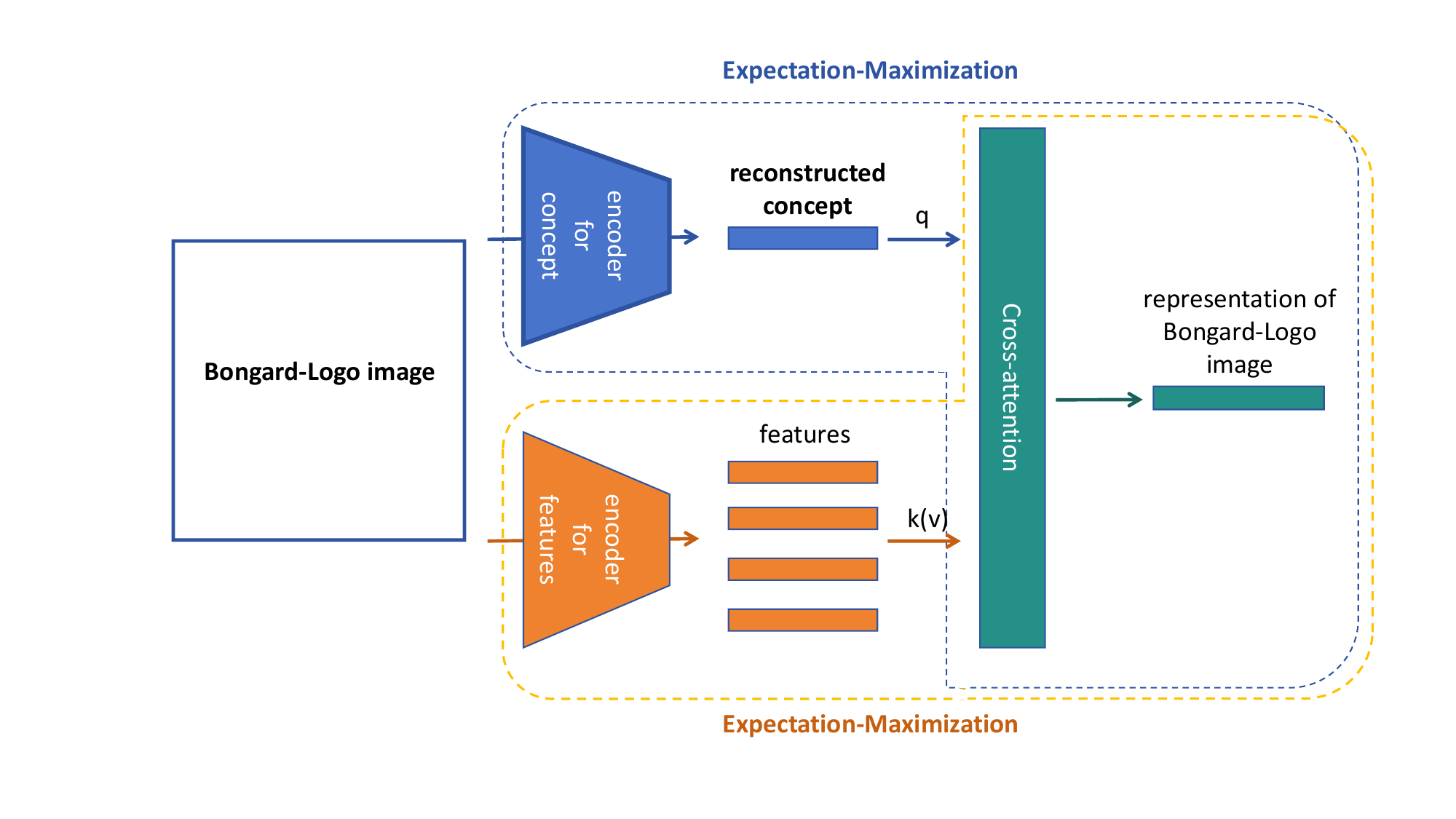}
	\caption{The position of the EM process incorporated into our framework}
\label{Framework and EM}
\end{figure}

%
The position of the EM process incorporated within our CFN is illustrated in Figure \ref{Framework and EM}. When the CFN's training process incorporates the EM process, we alternately optimize the modules outlined by the blue and yellow dashed lines in the figure. This approach allows us to integrate the search for the ideal concept set \( \{Y_\alpha\} \) into the CFN's optimization objectives.

In the CFN, the EM process refers to the mutually exclusive optimization of the parameters within the two modules $g_\omega(q_{ij} | x_{ij})$ and $g_\theta(k_{ij} | x_{ij})$.
\begin{enumerate}
    \item When optimizing the CFN, if we cease optimizing $g_\theta(k_{ij} | x_{ij})$, it is equivalent to finding a set of concepts $q_{ij}$ that can maximize the solution to the problem given the features $k_{ij}$. In other words, during the normal optimization process of the CFN, once we stop optimizing $g_\theta(k_{ij} | x_{ij})$, the optimization goal of the CFN shifts from maximizing the conditional distribution $P(z_{ij}, q_{ij}, k_{ij} | x_{ij})$ to maximizing the conditional distribution $P(z_{ij}, q_{ij} | x_{ij}, k_{ij})$. This latter form, which approximates the lower bound of network parameters given the latent variables, is the classic manifestation of the EM algorithm.

    \item If we stop optimizing $g_\omega(q_{ij} | x_{ij})$ and instead allow the optimization of $g_\theta(k_{ij} | x_{ij})$ to proceed, it is to enhance the capability of $g_\theta(k_{ij} | x_{ij})$ to extract features from $x_{ij}$.
\end{enumerate}

When we stop updating $g_\omega(q_{ij} | x_{ij})$ and allow the optimization of $g_\theta(k_{ij} | x_{ij})$ to continue, this is equivalent to running an EM algorithm where $k_{ij}$ are treated as the given latent variables. Conversely, if we do the opposite, it would be equivalent to running an EM algorithm where $q_{ij}$ are treated as the given latent variables. This paper refers to this process as the `dual EM algorithm.'

\subsection{Covariance-constrained Cross Feature Net (Triple-CFN)}

This paper posits that the current CFN, integrated with the dual EM algorithm, still has significant limitations. These limitations stem from the dual EM algorithm itself.

Put succinctly, when we cease the optimization of $g_\theta(k_{ij}|x_{ij})$ in CFN and instead release the optimization of $g_\omega(q_{ij}|x_{ij})$, it is akin to synthesizing a new set of concepts for the training instances of Bongard-Logo. These new concepts, while inexplicable and elusive to human understanding, cater well to the features of the given instances. Similarly, when reversing the optimized modules, our goal is to refine the feature extraction capability of $g_\theta(k_{ij}|x_{ij})$, hoping that the extracted features will better align with the given concepts. However, it is noteworthy that the features extracted by \( g_\theta(k_{ij}|x_{ij}) \) are tailored to the concepts re-synthesized within the training set, which may be markedly different from the original four concepts of Bongard-Logo. Therefore, these concepts are at risk of being overfitted to the training data. Allowing \( k_{ij} \) to cater to such concepts could also lead to excessive overfitting of \( g_\theta(k_{ij}|x_{ij}) \) to the training data, resulting in the failure of the CFN in generalization tasks. To overcome these limitations, the following methods have been devised in this paper.

\begin{figure}[htbp]\centering
	\includegraphics[trim=3cm 5cm 5cm 1cm, clip, width=8cm]{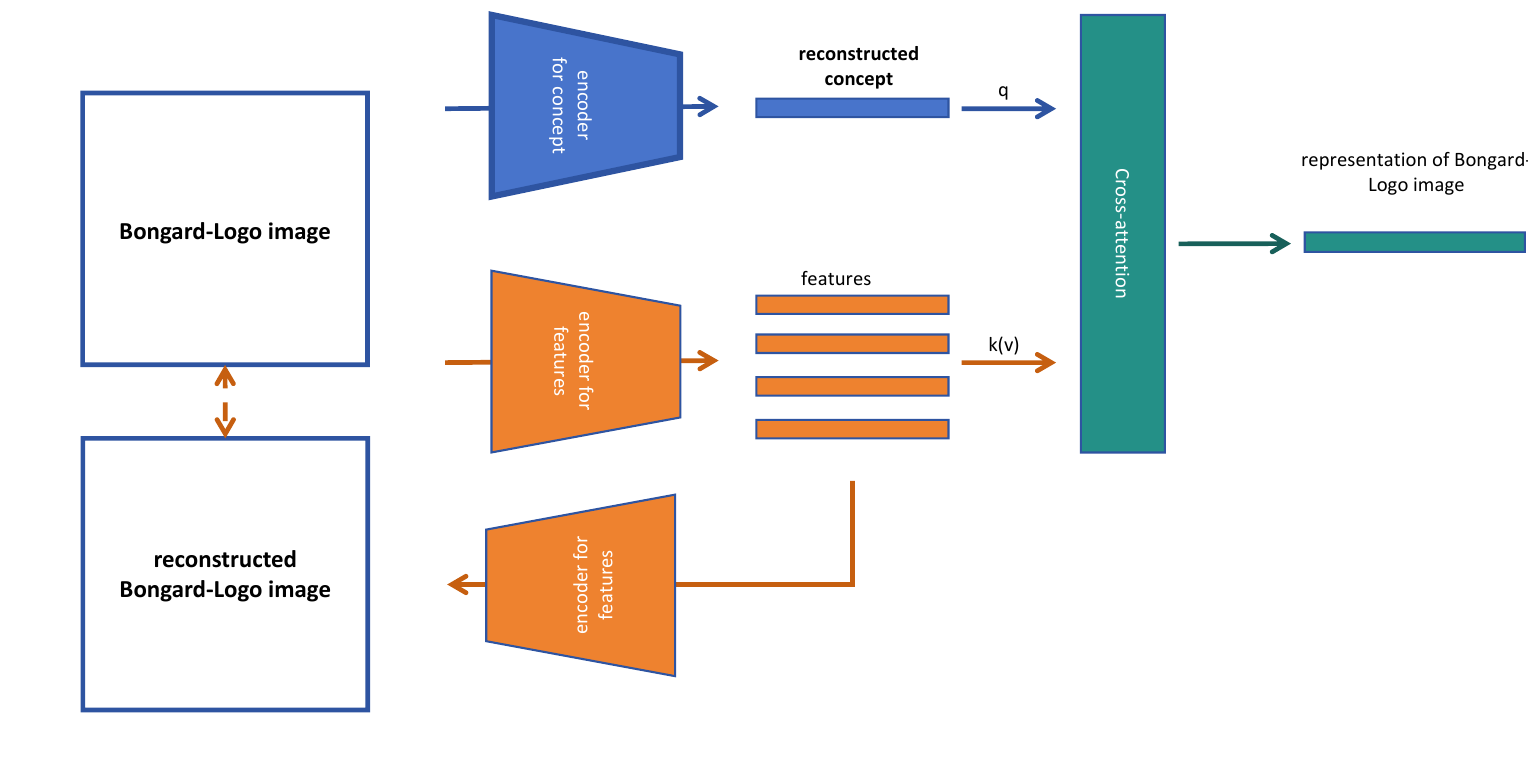}
	\caption{The CFN equipped with the capability for mutual information supervision}
\label{Triple-CFN-regression}
\end{figure}

If the features $k_{ij}$ extracted by $g_\theta(k_{ij} | x_{ij})$ are overly focused on the concepts $q_{ij}$ re-synthesized from the training data, then CFN will struggle to achieve excellent results in the generalization tasks of Bongard-Logo, such as the NV and CM concepts. Therefore, the paper calls for the design of an approach that allows the features \( k_{ij} \) not only to adapt to the concepts $q_{ij}$ but also to combat overfitting to a certain extent. 
Our objective is clear: to ensure that \( k_{ij} \) contains richer information regarding the Bongard-Logo images \( x_{ij} \), so as to enable the feature \( k_{ij} \) to cater more stably to the concept \( q_{ij} \), rather than overly accommodating it. 
To achieve this objective, the method proposed in this paper is to maximize the mutual information between $k_{ij}$ and $x_{ij}$.
The mutual information between  $x_{ij}$  and $ k_{ij}$ can be expressed as:
\begin{align}
I(X; K) = \sum_{x_{ij} \in X} \sum_{k_{ij} \in K} p(k_{ij}) p(x_{ij}|k_{ij}) \log \frac{p(x_{ij}|k_{ij})}{p(x_{ij})} 
\end{align}
From the formula, it can be seen that this mutual information supervision task can be achieved by setting up an extra regression task from $k_{ij}$ to $x_{ij}$. The CFN, augmented with the regression task, is depicted in the Figure \ref{Triple-CFN-regression}.

Unfortunately, the additional regression task introduces a significant computational cost, as the number of pixels in each Bongard-Logo image, $x_{ij}$, is considerable. Therefore, this paper had to abandon this approach and seek alternatives. Fortunately, an alternative method with satisfactory results was found, which is to impose decorrelation supervision on $k_{ij}$. This paper posits that, since our fundamental objective is to ensure that \( k_{ij} \) contains richer information, imposing decorrelation supervision among the dimensions of the latent representation \( k_{ij} \) can also fulfill this purpose.

\begin{figure}[htp]\centering
	\includegraphics[trim=2cm 0cm 6cm 0cm, clip, width=8cm]{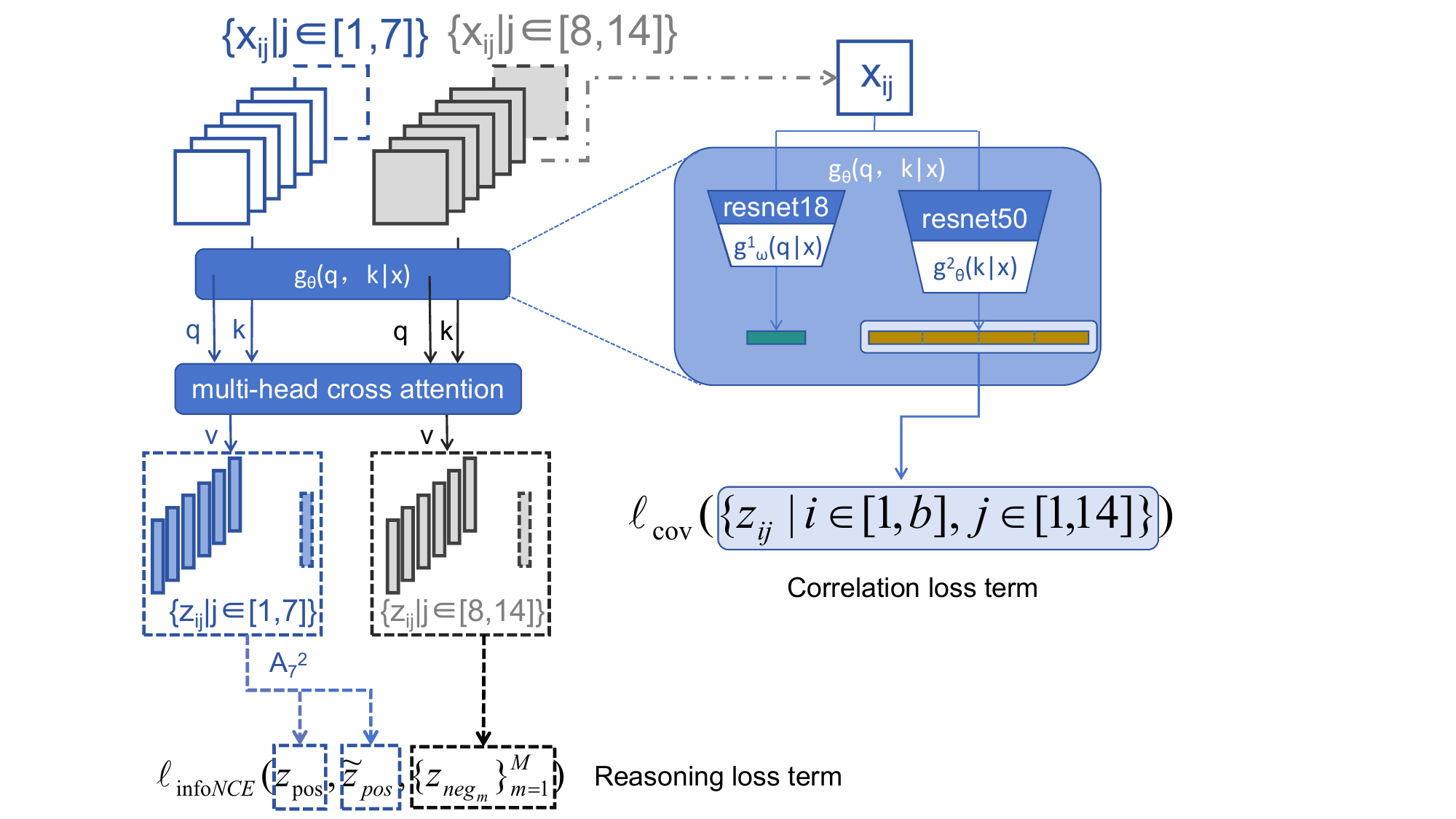}
	\caption{Structure of Triple-CFN}
\label{Triple-CFN}
\end{figure}

To achieve a decorrelation approach, we introduced an additional loss term based on the covariance matrix for CFN to decrease the correlation between each dimension in $k_{ij}$. 
the covariance matrix of each element in $k_{ij}$ is calculated using Formula (\ref{matrix_cov}). The additonal correlation loss for CFN is calculated using Formula (\ref{eq_cov}).
This resulted in the creation of the Covariance-constrained Cross Feature Net (Triple-CFN). 
Figure \ref{Triple-CFN} presents a detailed depiction of the feed-forward process within Triple-CFN.

\subsection{Triple-CFN on RPM problem}

When dealing with RPM problems, we firmly believe that Triple-CFN can also demonstrate its distinctive utility and value. Because this paper concerns that the methodology of separately extracting concepts and features in CFN, the method of using the dual EM algorithm to resynthesize non-conflicting concepts for training data, and the strategy of reducing overfitting in the feature extraction process through decorrelation supervision are all key to tackling the problem of visual abstract reasoning.

The RPM problem consists of a problem statement with a missing piece and an option pool. 
When the missing piece is filled with the correct option, the completed $3\times3$ matrix exhibits a self-consistent progressive pattern, whereas incorrect options do not.
An RPM solver needs to understand the progressive pattern in the problem statement and identify the option that follow this pattern. 

\begin{figure}[htp]\centering
	\includegraphics[trim=0cm 0cm 0cm 0cm, clip, width=8cm]{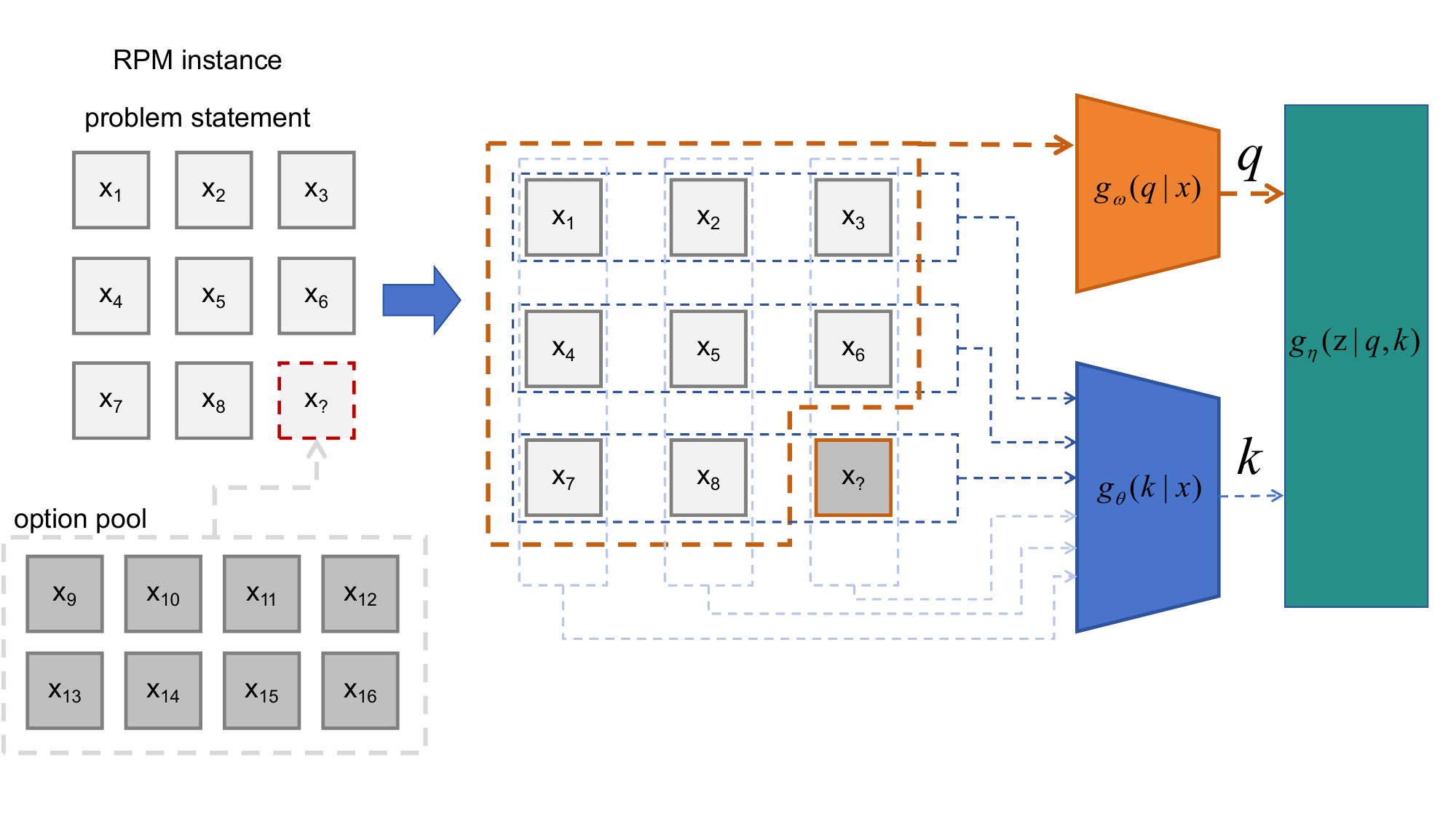}
	\caption{CFN framework for adapting to RPM problems.}
\label{CFN framework for adapting to RPM problems}
\end{figure}

In light of the reasoning characteristics of RPM, this paper proposes to follow the pattern shown in Figure \ref{CFN framework for adapting to RPM problems}, utilizing CFN to extract the concept $q$ from the problem statement and the row-column progressive features $k$ from the completed $3\times 3$ matrix. On this basis, the Triple-CFN mechanism and the dual EM process can be further applied.
Therefore, we suggest that when addressing RPM problems, the core structure of Triple-CFN remains unchanged, with only adjustments to $g_\omega(q|x)$ and $g_\theta(k|x)$ by replacing their backbone networks, aiming to adapt it to RPM problems with smaller images and a larger volume of instances.

\begin{figure}[htp]\centering
	\includegraphics[trim=2cm 0cm 8cm 0cm, clip, width=7cm]{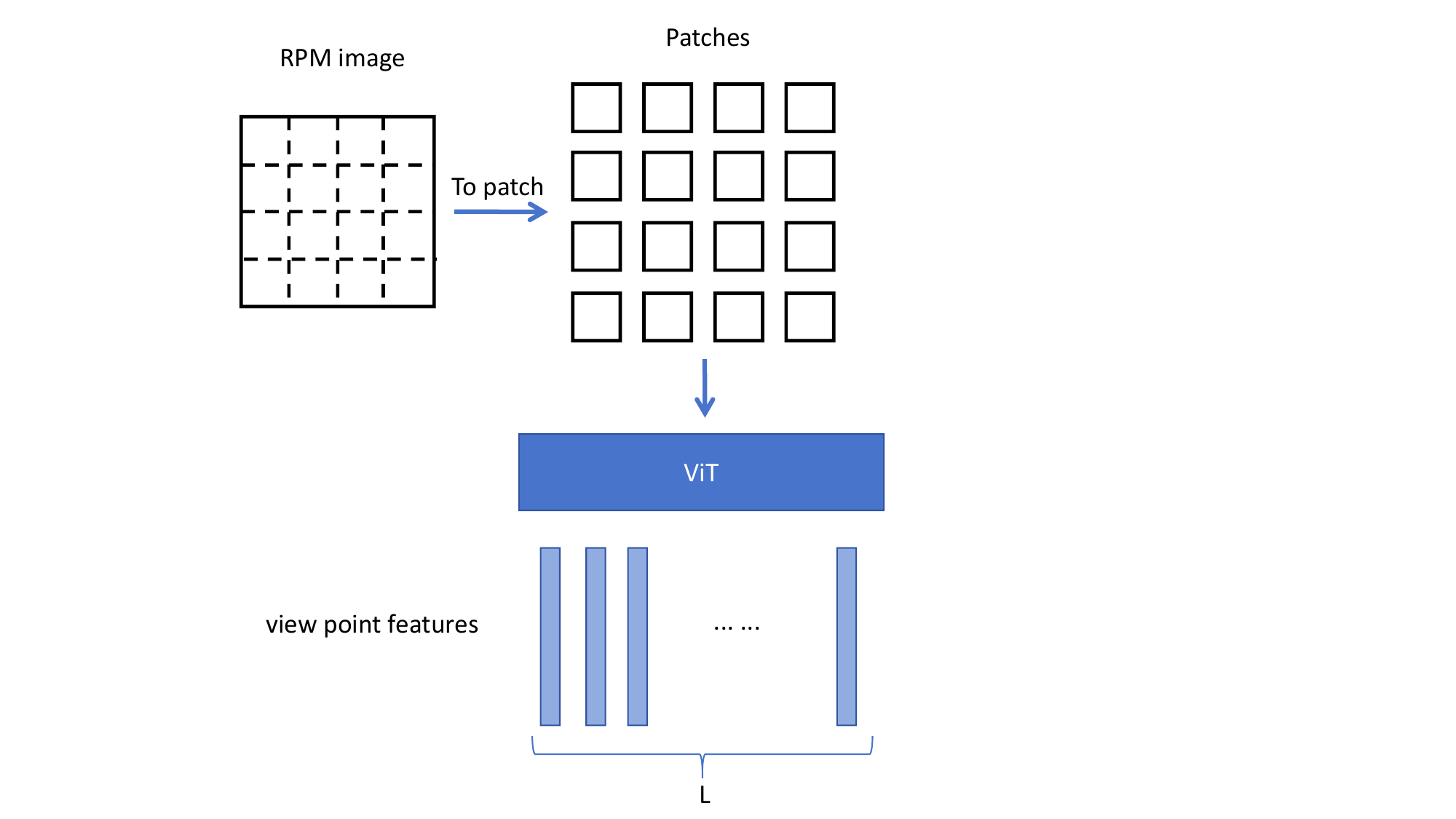}
	\caption{Structure of Triple-CFN's perceptron}
\label{perceptron Triple-CFN}
\end{figure}

Following the current consensus among RPM solvers, which emphasizes the need for multi-scale or multi-viewpoint feature extraction from RAVEN images\cite{SAVIR-T,RS,SCL}, Triple-CFN utilizes Vision Transformer (ViT)\cite{ViT, A survey of visual transformers} for feature extraction while preserving all output vectors as multi-viewpoint features. In this paper, the number of viewpoints is denoted as $L$. The process of the extraction can be expressed in figure \ref{perceptron Triple-CFN}.
Subsequently, Triple-CFN processes each viewpoint equally.

\begin{figure}[htp]\centering
	\includegraphics[trim=0cm 0cm 12cm 0.5cm, clip, width=7cm]{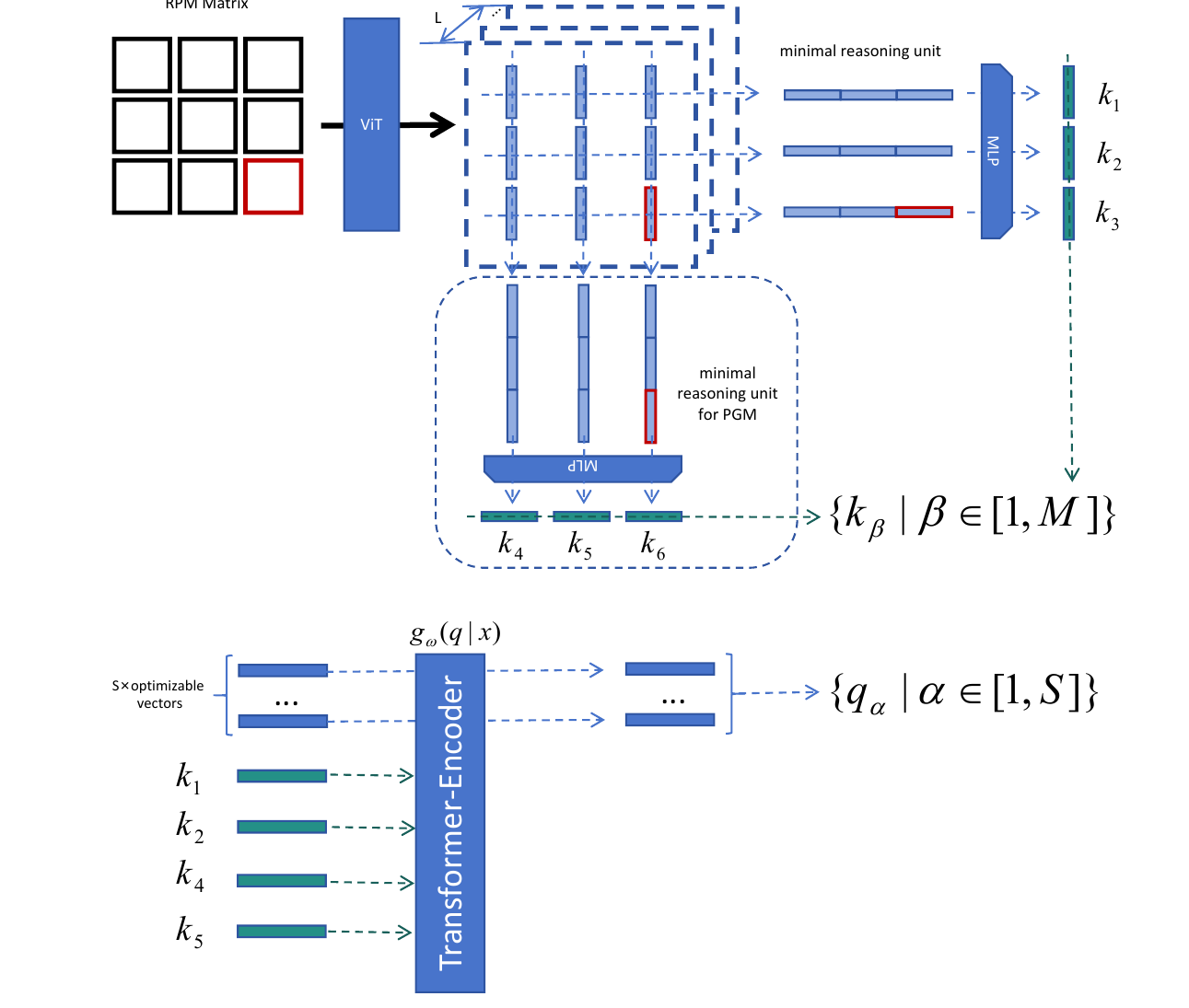}
	\caption{The calculation process of $g(q, k|\,x)$ in Triple-CFN}
\label{calculation Triple-CFN}
\end{figure}

\begin{figure}[htp]\centering
	\includegraphics[trim=0cm 0cm 3cm 1cm, clip, width=6.5cm]{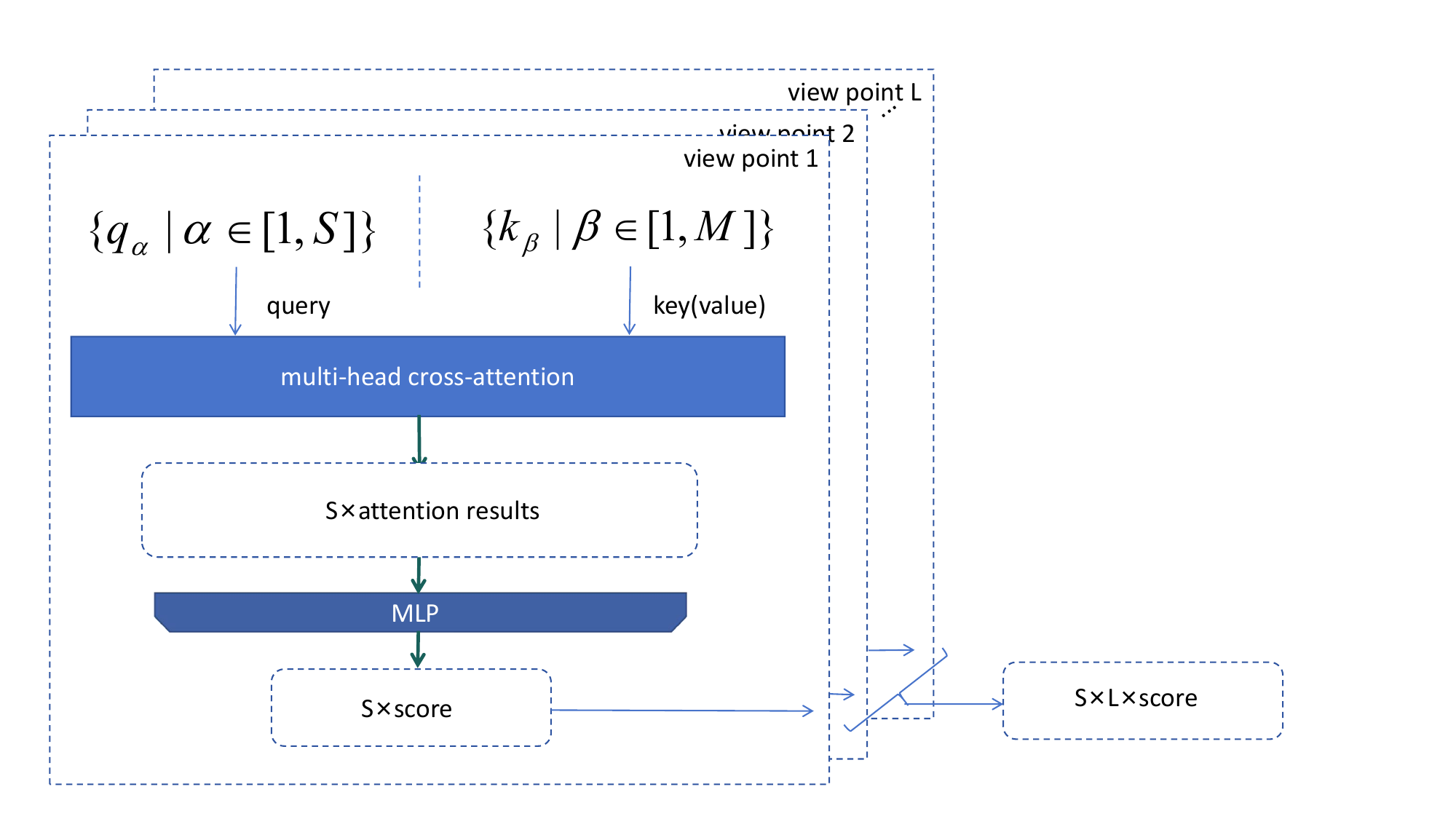}
	\caption{The calculation process of final score in Triple-CFN}
\label{The calculation process offinal score in Triple-CFN}
\end{figure}

Subsequently, we designed the new structures for \( g_\omega(q|x) \) and \( g_\theta(k|x) \) as illustrated in the figure \ref{calculation Triple-CFN}.
The upper half of the figure illustrates the process by which Triple-CFN extracts the features \( k_\beta \) of RPM instances, while the lower half depicts the process of extracting the concepts \( q_\alpha \).
\begin{enumerate}
    \item In the upper half of Figure \ref{calculation Triple-CFN}, it can be observed that Triple-CFN utilizes a Multi-layer Perceptron (MLP) with a bottleneck structure to extract progressive information from both the row and column directions of the $3 \times 3$ multi-viewpoint feature matrix for each viewpoint, generating feature vectors $\{k_\beta | \beta \in [1,M]\}_l$. Here, $l$ represents the identifier for the viewpoint, and $l \in [1, L]$. This MLP, along with the previously mentioned ViT, constitutes the new structure for $g_\omega(k | x)$.
    
    \item In the lower half of Figure \ref{calculation Triple-CFN}, it is shown that Triple-CFN combines the feature vectors $k_\beta$ extracted from the problem statement with $S$ optimizable vectors, and inputs them into the network $g_\omega(q|x)$, which is underpinned by a Transformer-Encoder. After processing, it captures all optimizable vectors from the output of $g_\omega(q|x)$ to compute the concept vectors $\{q_\alpha | \alpha \in [1, S]\}_l$. The hyperparameter $S$ determines the number of concept vectors extracted for the RPM problem statement. In this paper, it is designated as 2. However, the implications of assigning a larger value to $S$ on the performance of the Triple-CFN have not been investigated within the scope of this study.
\end{enumerate}
Given $\{q_\alpha\}_l$ and $\{k_\beta\}_l$, the $g_\eta(z|q,k)$ module of Triple-CFN can fulfill its role, where $g_\eta(z|q,k)$ utilizes the Transformer-Decoder as its backbone. After calculating the cross-attention results between $\{q_\alpha\}_l$ and $\{k_\beta\}_l$, a new MLP is used to score these outputs, resulting in $L\times S$ logical scores. Subsequently, the sets of $L\times S$ logical scores are averaged to determine the final score. The calculation process of the final score can be expressed in Figure \ref{The calculation process offinal score in Triple-CFN}.

To optimize this RPM-version Triple-CFN, we impose constraints on this final score by employing the Cross-Entropy loss function as a reasoning loss term. Additionally, we apply the correlation loss term on the feature vector set $\{k_\beta | \beta \in [1, M]\}_l$.

\subsection{Summary for the Framework}
The CFN framework advocates for separate extraction of concepts and features, and on this basis, this paper further proposes that concept extraction should closely align with the reasoning features of the problem. However, it is emphasized that feature extraction should not blindly adhere to synthetic concepts in the training set; instead, it should retain more information conducive to reasoning. To this end, this paper skillfully integrates a dual EM process and a decorrelation supervision mechanism. It is worth noting that neither the EM process nor decorrelation supervision is new technology, nor are they being applied to the field of deep learning for the first time. The contribution of this paper is to identify the key challenges of visual abstract reasoning problems, highlight the essential traits of a sound reasoning problem solver, and purposely integrate the EM process and decorrelation supervision into the solver's design based on these trait requirements.

\section{Meta Covariance-constrained Cross Feature Net (Meta Triple-CFN)}


In the RPM problem, in addition to the label of the correct answer, each instance is accompanied by a clear description of its progressive pattern, presented in the form of metadata. This metadata serves as an excellent record of the concepts involved in the reasoning instances, aligning well with the exploration pursuits of the dual EM process. Therefore, this paper proposes that metadata can be utilized as additional supervisory information to guide the extraction process of concepts ${q_\alpha}$ in Triple-CFN, potentially achieving better results in terms of interpretability and reasoning accuracy compared to the dual EM algorithm.

\subsection{Metadata-Supervised EM Process
}

Previous works have attempted to enhance RPM solvers by incorporating additional tasks of learning image progression patterns. The aim is to improve the solvers' performance on reasoning tasks and enhance their interpretability. These extra pattern-matching tasks typically utilize the metadata provided by RPM problems as supervisory signals. However, research led by MRNet \cite{MRNet} suggests that this approach may be counterproductive.

This paper argues that Triple-CFN, which explicitly separates the processes of feature extraction and concept extraction, offers a different path. By supervising only the concept extraction process using metadata, without interfering with the feature extraction process, it may be possible to reverse the phenomenon where increased supervisory signals impair the solver's reasoning accuracy\cite{MRNet}. Previous RPM solvers, such as RS-Tran \cite{RS} and MRNet, did not separate these two processes, and therefore this was an approach they had not explored.

\begin{figure}[htp]\centering
	\includegraphics[trim=0cm 5cm 0cm 0cm, clip, width=8.5cm]{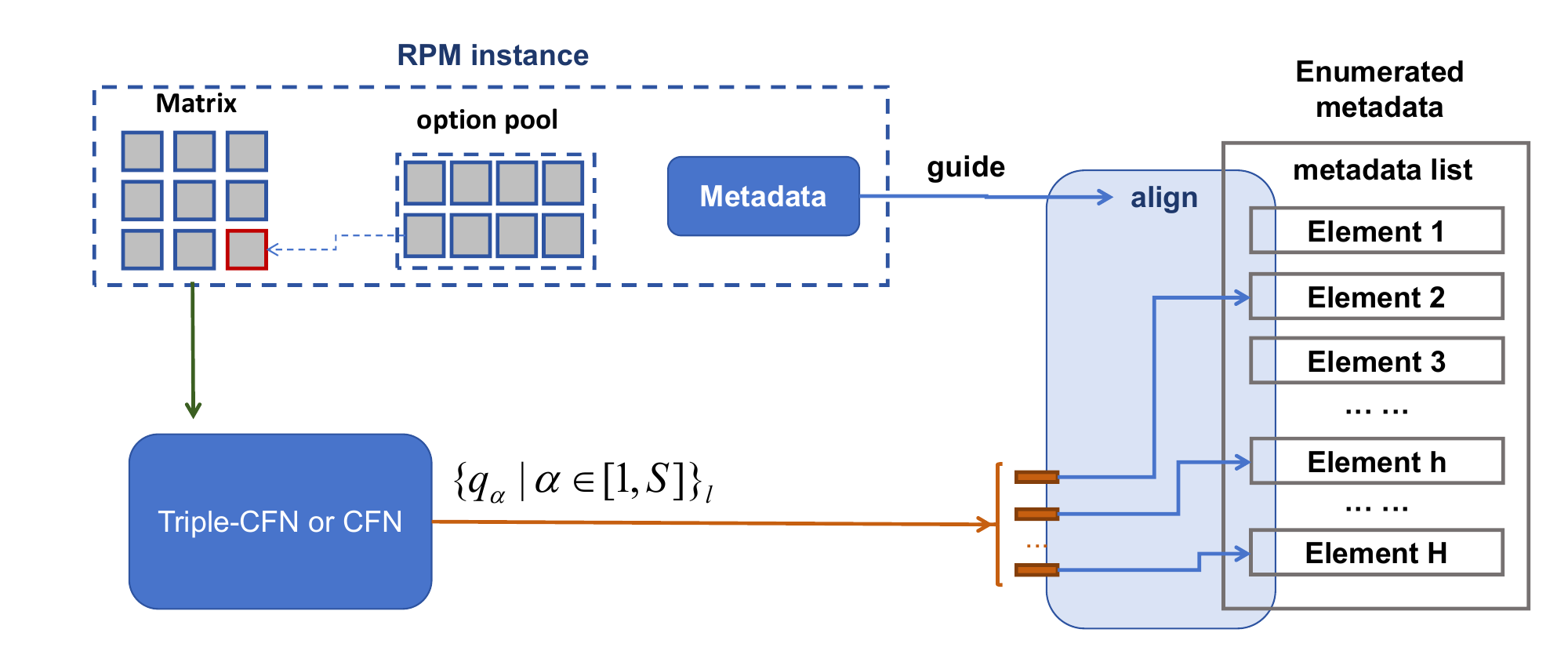}
	\caption{The utilization of metadata}
\label{Meta_Triple_CFN_abstruct}
\end{figure}

The method of utilizing metadata adopted in this paper differs from previous works. This paper has had successful experiences in enhancing the concept extraction capability in (Triple-)CFN using the EM algorithm, thus this paper has designed the metadata utilization approach as shown in the Figure \ref{Meta_Triple_CFN_abstruct}.
\begin{enumerate}
    \item In the figure, on the right side, we can see that we first enumerate the metadata of all instances in the RPM database (such as RAVEN, PGM) to generate a metadata list.
    \item  As shown on the left side of the figure, when Triple-CFN processes an instance from the RPM database, we align the concepts $\{q_\alpha\}_l$ extracted by Triple-CFN with the corresponding elements in the newly created metadata list according to the instance's metadata, thereby directly providing supervisory signals for the extraction of concepts. This is similar to conducting supervised learning for the E-step in the EM algorithm, and therefore it can be said that this meta-process is in line with the dual EM process.
\end{enumerate}

In this way, we no longer need to utilize the EM algorithm to estimate the expected values of concepts for given features; instead, we can directly align the concepts to a well-defined superior concept space, which also avoids the local optimum problem in the EM algorithm to some extent.


\subsection{The Structure of Meta Triple-CFN}
The specific details of the utilization strategy for metadata are as follows: Considering the necessity to impose constraints on the $\{q_\alpha | \alpha \in [1, S]\}$ across all viewpoints, this paper opts to compute the average of $\{q_\alpha | \alpha \in [1, S]\}$ for $l \in [1, L]$, denoted as $\{\overline{q}_\alpha | \alpha \in [1, S]\}$, and to apply constraints to this averaged representation. Subsequently, we enumerate all metadata in the RPM database to form a metadata list. We denote the total number of elements in this list as $K$. Furthermore, since the metadata contains descriptions of the progressive patterns corresponding to the instances, we process the $K$ elements in the metadata list using a standard Transformer-Encoder, thus forming a code book composed of $K$ vectors. This code book is denoted as $\{T_k | k \in [1, K]\}$.

Finally, we leverage the InfoNCE loss function once more to impose additional constraints on $\{\overline{q}_\alpha|\, \alpha\in [1, S]\}$. The vectors within $\{\overline{q}_\alpha|\, \alpha\in [1, S]\}$ are aligned with the corresponding positions in the code book \( \{ T_k | k \in [1, K] \} \) as dictated by the metadata. This constraint can be represented by the following formula: 
\begin{align}
    {\ell _\text{Metadata}}= -\sum_{\alpha=1}^{\tilde{S}}  \sum_{\tilde k = 1}^{K} y_{{\alpha\tilde k}}\cdot\log \frac{{{e^{({\bar q_{\alpha}} \cdot {  T_{\tilde k}})/\tau}}}}{\sum_{k = 1}^{K} {{e^{({\bar q_{\alpha} } \cdot {T_{k}})/\tau}}} }\label{act3}
\end{align}
This formula represents a standard form for constraining cosine similarity between vectors using cross-entropy loss. Here, $y_{{\alpha\tilde{k}}}$ indicates whether the content of the $\alpha$-th progressive pattern recorded in the instance metadata matches the $\tilde{k}$-th element in the codebook. The temperature coefficient $\tau$ is $10^{-6}$, and $\tilde{S} = S - 1$. In the formula, it can also be observed that we only impose constraints on $\tilde{S}$ vectors within the set $\{\overline{q}_\alpha | \alpha \in [1, S]\}$, leaving one vector free. Here, $\tilde{S}$ refers to the number of carriers of the progressive patterns present in the RPM instance, which also corresponds to the number of progressive patterns recorded in the metadata. This implies that when we apply constraints to $\{\overline{q}_\alpha | \alpha \in [1, S]\}$ using metadata, the size of $S$ can no longer be arbitrarily set but must be configured larger than $\tilde{S}$. The free vector serves as a safeguard against certain subtle and unreasonable configurations within the metadata that might be unforeseen.

\begin{figure}[htp]\centering
	\includegraphics[trim=0cm 0cm 0cm 0cm, clip, width=8.5cm]{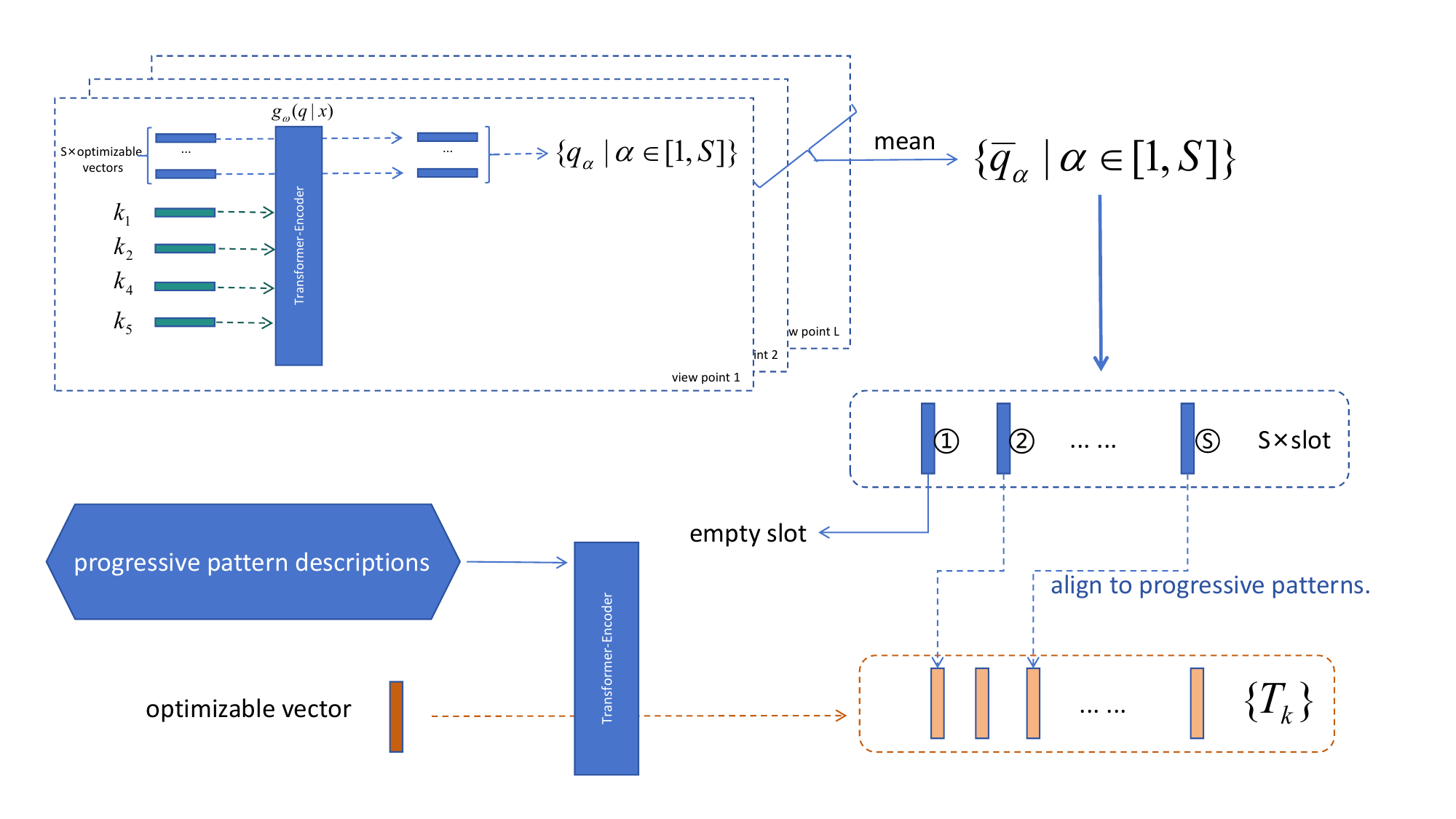}
	\caption{The calculation process of $\ell _\text{Meta}$ in Triple-CFN}
\label{Meta}
\end{figure}

The detailed process of imposing constraints on \( q_\alpha \) using metadata can be depicted as shown in the Figure \ref{Meta}.
The Triple-CFN, augmented with such constraints, is named Meta Triple-CFN in this paper.


\section{Re-space layer}
The core of the Meta-process lies in constructing a concept space $\{T_k | k \in [1, K]\}$ for Triple-CFN. By skillfully aligning the concept vectors $\{q_\alpha\}$ extracted by Triple-CFN with the corresponding positions in this concept space, the performance of Triple-CFN is significantly enhanced, and its interpretability is greatly improved. Correspondingly, this paper also constructs a similar space for the feature extraction process, aiming to bring deeper breakthroughs and innovations to Triple-CFN.

This paper proposes an innovative layer structure—the Re-space Layer, which serves as a special network that constructs an exclusive vector space for the feature $k_\beta$. By utilizing this space to normalize the feature vectors $k_\beta$, it enables $k_\beta$ to adapt more stably to the concepts.

\subsection{The Structure of Re-space Layer}
\begin{figure}[htp]\centering
	\includegraphics[trim=2cm 5cm 2cm 3.5cm, clip, width=8.5cm]{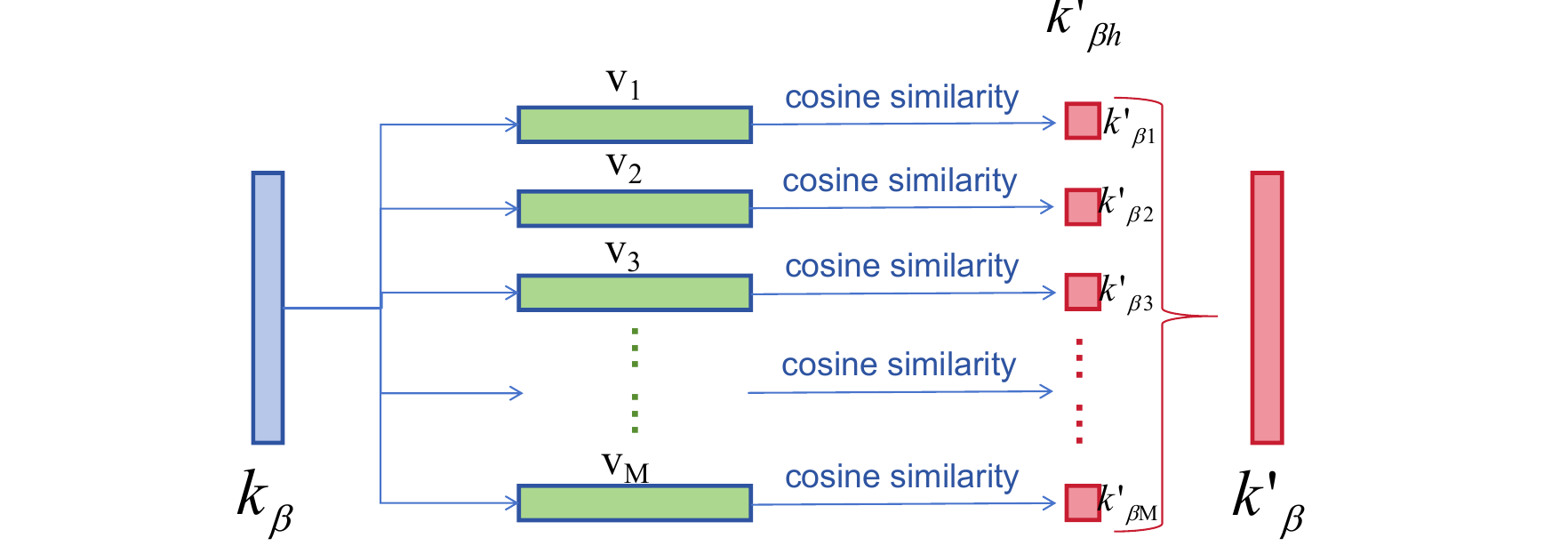}
	\caption{Structure of Re-space layer}
\label{respace}
\end{figure}

Specifically, we establish $M$ optimizable vectors for Triple-CFN, which depict a vector space $\{v_h|\,h\in [1,M]\}$. The hyperparameter $M$ is set to be as large as the dimension of $k_\beta$. Subsequently, cosine similarity is computed between the features $\{k_\beta\}$, and each $M$ optimizable vector. And the calculate process can be expressed as follows:

\begin{align}
     k'_{\beta h} &= \frac{v_h \cdot k_\beta}{|\,|\,v_h|\,|\, \times|\,|\,k_\beta|\,|\,}\\
     k'_\beta  &= \{k'_{\beta h}|\, h\in [1,M]\}
\end{align}
The computed vector \( k'_\beta \), constituted by \( M \) cosine similarities \( \{k'_{\beta h} | h \in [1, M]\} \), signifies the coordinates of each feature vector \( k_\beta \) within the vectorial space \( \{v_h | h \in [1, M]\} \). The original feature vectors \( \{k_\beta\} \) are superseded by the calculated coordinates \( \{k'_\beta \} \) for subsequent reasoning tasks. The procedure of the Re-space Layer is elucidated in Figure \ref{respace}.

During model training, the similarity among the $K$ optimizable vectors is constrained to ensure a richly diverse vector space and avoid collapse from the Re-space layer output. The constraint is implemented by utilizing the following function as an additional loss term for Triple-CFN or Meta Triple-CFN:

{\footnotesize\begin{align}
    &{\ell_\text{Re-space}}(\{v_h\}^M_{h=1})) =\sum_{h = 1}^M  -\log \frac{{{e^{({v_h} \cdot {v_h})/t}}}}{{{e^{({v_{h}} \cdot { v_{h}})/t}} + \sum_{\tilde{h}= 1,\, \tilde{h} \neq h}^M {{e^{({v_{h}} \cdot {v_{\tilde h}})/t}}} }}
\end{align}}

\noindent Where the $t$ is set to $10^{-2}$. 

\subsection{The Initialization Process of the Re-space Layer.
}
It is worth noting that integrating the Re-space layer with (Meta) Triple-CFN requires a warm-start process.
Specifically, before we integrate the Re-space layer into (Meta) Triple-CFN, we need to completely train a (Meta) Triple-CFN and retain the parameters of its $g_\theta(k|\,x)$ network as warm-start parameters. 

\subsection{The Dropping Process of the Re-space Layer.
}
In RPM problems, there also exist traditional generalization tasks, such as the Interpolation and Extrapolation datasets in the PGM problem \cite{PGMdataset}. Given our previous discussion that the EM process may weaken a model's generalization ability, Meta Triple-CFN, as an inheritor of the EM process, is no exception. Particularly, the metadata utilized in Meta Triple-CFN faces Out-Of-Distribution (OOD) situations in multiple generalization tasks of PGM. Therefore, the Re-space Layer explores the possibility of forgoing the use of metadata during its warm-start process to enhance the performance of Meta Triple-CFN on generalization tasks. Specifically, once the training of Meta Triple-CFN is complete, we not only retain the parameters of $g_\theta(k|x)$ as warm-start parameters for the Re-space Layer, but also discard the constraints imposed by the enumerated meta list on the concept set $\{q_\alpha\}_l$. Evidently, this paper aims to leverage ``catastrophic forgetting'' during the warm-start process of the Re-space Layer to counteract the issue of features excessively catering to concepts.

\section{Experiment}

All our experiments are implemented in Python using the PyTorch\cite{
Pytorch} framework. Additionally, our experiments were conducted on a single server equipped with four A100s graphics processing units.

\subsection{Experiment on Bongard-Logo}
In this study, we conducted experiments on the Bongard-Logo dataset using the designed CFN and Triple-CFN models. To demonstrate the impact of the dual EM process on model performance, we performed ablation experiments. We trained the models using mini-batch gradient descent with a batch size of 120. During training, we utilized the Adam optimizer \cite{ADAM} with a learning rate of $10^{-3}$ and a weight decay of $10^{-4}$. The Adam optimizer is a commonly used method for deep network optimization. The results are presented in Table \ref{Bongard_Results}.

\begin{table}[h]
\caption{Reasoning Accuracies of CFN on Bongard-logo.}
\label{Bongard_Results}
\centering
\begin{tabular}{ccccccc}
\toprule
&\multicolumn{5}{c}{ Accuracy(\%)}& \\
\cmidrule{2-6}
Model&Train& FF&BA&CM&NV \\
\midrule
SNAIL&59.2&56.3&60.2&60.1&61.3\\
\midrule
ProtoNet&73.3&64.6&72.4&62.4&65.4\\
\midrule
MetaOptNet &75.9&	60.3&	71.6&	65.9&	67.5\\
\midrule
ANIL & 69.7 & 56.6 & 59.0 & 59.6 & 61.0\\
\midrule
Meta-Baseline-SC & 75.4 & 66.3 & 73.3 & 63.5 & 63.9 \\
\midrule
Meta-Baseline-MoCo & 81.2 & 65.9 & 72.2 & 63.9& 64.7 \\
\midrule
WReN-Bongard & 78.7 & 50.1 & 50.9 & 53.8 & 54.3 \\
\midrule
SBSD&83.7&75.2&91.5&71.0&74.1\\
\midrule
PMoC&92.0&90.6&97.7&77.3&76.0\\
\midrule
\midrule
PMoC+Re-space Layer&93.0&91.2&98.1&77.5&76.2\\
\midrule
CFN&92.2&88.5&98.1&77.0&77.3\\
\midrule
CFN+EM&\underline{93.6}&\underline{92.3}&\underline{98.3}&77.2&77.0\\
\midrule
Triple-CFN&93.2&92.0&{98.2}&\underline{78.0}&\underline{78.1}\\
\midrule
Triple-CFN+EM&\textbf{94.0}&\textbf{92.6}&\textbf{98.5}&\textbf{78.3}&\textbf{78.5}\\
\bottomrule
\end{tabular}
\end{table}

As observed in Table \ref{Bongard_Results}, the dual EM process enhanced the CFN's performance on the FF and BA problems without significantly affecting its ability to solve the generalization problems of NV and CM. However, Triple-CFN mitigates this limitation to a certain extent, representing a promising contribution. Unsurprisingly, the integration of Triple-CFN with the EM process achieved the most optimal reasoning performance. It is worth noting that we did not conduct experiments on the combination of Triple-CFN and the Re-space Layer for the Bongard-Logo problem. The reason is that the Re-space Layer compresses its output values into the range of $[-1,1]$, which would adversely affect the optimization of Triple-CFN when using the InfoNCE loss function. To verify the effectiveness of the Re-space Layer on the Bongard-Logo problem, we turned our attention to the current state-of-the-art model, PMoC \cite{PMoC}, which uses cross entropy as its loss function. We integrated the Re-space Layer into the feature extraction process of PMoC and conducted corresponding experiments.

Finally, we notice the augmentation experiments conducted on PMoC\cite{PMoC}, which aimed to expand the Bongard-Logo database through data augmentation techniques, thereby observing the potential and reasoning capacity of the model \cite{PMoC}. 
The data augmentation methods employed include random rotations (90$^\circ$, 180$^\circ$, and 270$^\circ$) and (horizontal or vertical) flips \cite{PMoC}. 
These augmentation methods do not alter the concept composition within Bongard-Logo but rather aim to diversify the expressions of concepts\cite{PMoC}. 
This paper posits that enriching the expressions of concepts at pixel level can unlock further potentials of the network, enabling deeper exploration of the network's capacity to accommodate, compress, extract, and abstract reasoning patterns. This is one of the original intentions behind the establishment of the Bongard-Logo database\cite{Bongard2}.
Therefore, this paper also revisited the Triple-CFN experiments under the same augmentation approach. The experimental results are presented in the Table \ref{augmented_Bongard_Results}.

\begin{table}[h]
\caption{Reasoning Accuracies of CFN on Augmented Bongard-logo.}
\label{augmented_Bongard_Results}
\centering
\resizebox{\linewidth}{!}{
\begin{tabular}{ccccccc}
\toprule
&\multicolumn{5}{c}{ Accuracy(\%)}& \\
\cmidrule{2-6}
Model&Train& FF&BA&CM&NV \\
\midrule
PMoC&94.5&92.6&98.0&78.3&76.5\\
\midrule
\midrule
PMoC+Re-space Layer&95.0&93.3&99.0&80.1&79.8\\
\midrule
CFN&94.7&91.5&98.8&78.5&78.3\\
\midrule
CFN+EM&{94.9}&{93.8}&{99.4}&78.8&78.2\\
\midrule
Triple-CFN&94.9&93.0&{99.2}&{80.8}&{79.1}\\
\midrule
Triple-CFN+EM&\textbf{95.3}&\textbf{94.3}&\textbf{99.8}&\textbf{81.3}&\textbf{82.0}\\
\bottomrule
\end{tabular}
}
\end{table}

The table \ref{augmented_Bongard_Results} demonstrates that the augmented Bongard-Logo has not overshadowed the superiority of Triple-CFN compared to PMoC\cite{PMoC}.
Additionally, considering Triple-CFN's lower computational overhead and lighter parameter configuration compared to PMoC, we posit that its ability to achieve better results on more abundant and augmented data is promising. It is noteworthy that the Re-space Layer brings a more significant improvement to the PMoC model on the augmented Bongard-Logo dataset, further demonstrating its value on a larger scale of data.

\subsection{Experiment on RPM}

\subsubsection{RAVEN} When confronted with the RAVEN database, which is a type of RPM problem, both Triple-CFN and its combination with the Re-space layer have demonstrated considerable performance. In this study, we conducted experiments using the same software and hardware configurations as those employed in RS-Tran \cite{RS}. We replicated the experimental parameters from the RS-Tran setup, including batch size, learning rate, data volume, and all other factors that could potentially influence model performance. This was done to ensure the fairest comparison with RS-Tran, which is currently considered the state-of-the-art model. The accuracy of Triple-CFN on RAVEN and I-RAVEN is recorded in Table \ref{RAVEN_IRAVEN_Results}. The results indicate that Triple-CFN exhibits promising performance. It is worth noting that Triple-CFN has almost half the number of parameters compared to RS-Tran.

\begin{table}[h]
\caption{Reasoning Accuracies on RAVEN and I-RAVEN.}
\label{RAVEN_IRAVEN_Results}
\centering
\resizebox{\linewidth}{!}{
\begin{tabular}{cccccccccc}
\toprule
\toprule
&\multicolumn{8}{c}{Test Accuracy(\%)}& \\
\cmidrule{2-9}
Model&Average&Center&2 $\times$ 2 Grid&3 $\times$ 3 Grid&L-R&U-D&O-IC&O-IG \\
\midrule
{CoPINet\cite{CoPINet}}&52.96/22.84&49.45/24.50&61.55/31.10&52.15/25.35&68.10/20.60&65.40/19.85&39.55/19.00&34.55/19.45 \\
\cmidrule{2-9}
{PrAE Learner\cite{PrAE}}&65.03/77.02&76.50/90.45&78.60/85.35&28.55/45.60&90.05/96.25&90.85/97.35&48.05/63.45&42.60/60.70 \\
\cmidrule{2-9}
SAVIR-T \cite{SAVIR-T}&94.0/98.1&97.8/99.5&94.7/98.1&83.8/93.8&97.8/99.6&98.2/99.1&97.6/99.5&88.0/97.2\\
\cmidrule{2-9}
SCL \cite{SCL, SAVIR-T}&91.6/95.0&98.1/99.0&91.0/96.2&82.5/89.5&96.8/97.9&96.5/97.1&96.0/97.6&80.1/87.7\\
\cmidrule{2-9}
MRNet \cite{MRNet}&96.6/-&-/-&-/-&-/-&-/-&-/-&-/-&-/-\\
\cmidrule{2-9}
RS-TRAN\cite{RS}&{98.4}/98.7&99.8/{100.0}&{99.7}/{99.3}&{95.4}/96.7&99.2/{100.0}&{99.4}/99.7&{99.9}/99.9&{95.4}/95.4 \\
\cmidrule{2-9}
Triple-CFN&98.9/{99.1}&100.0/{100.0}&99.7/{99.8}&96.2/{97.5}&99.8/{99.9}&99.8/{99.9}&99.9/99.9&97.0/{97.3} \\
\cmidrule{2-9}
Triple-CFN+Re-space&99.4/\textbf{99.6}&100.0/\textbf{100.0}&99.7/\textbf{99.8}&98.0/\textbf{99.1}&99.9/\textbf{100.0}&99.9/\textbf{100.0}&99.9/\textbf{99.9}&98.5/\textbf{99.0} \\
\bottomrule
\bottomrule
\end{tabular}
}
\end{table}

\subsubsection{PGM} We conducted experiments on the PGM dataset under the same experimental conditions as Rs-Tran\cite{RS}, the accuracy of reasoning is recorded in the Table \ref{PGM_Results} and the accuracy of reasoning progressive patterns is recorded in Tabel \ref{PGM_Pattern_Results}. Given that the Meta Triple-CFN is equivalent to the Triple-CFN augmented with a supervised version of the EM process, we have not undertaken experiments on the Meta Triple-CFN with the additional application of the EM process.
These findings in Table \ref{PGM_Results} and \ref{PGM_Pattern_Results} are aimed at illustrating Meta Triple-CFN's capability to attain high reasoning accuracy while simultaneously maintaining the interpretability of progressive patterns. This is not achievable by other previous model \cite{MRNet,RS}.

\begin{table}[h]
\caption{Reasoning Accuracies of Triple-CFN on PGM.}
\label{PGM_Results}
\centering
\begin{tabular}{ccc}
\toprule
Model&Test Accuracy(\%) \\
\midrule
SAVIR-T \cite{SAVIR-T}&91.2\\
\midrule
SCL \cite{SCL, SAVIR-T}&88.9\\
\midrule
MRNet \cite{MRNet}&94.5\\
\midrule
RS-CNN\cite{RS}&82.8\\
\midrule
\midrule
RS-TRAN\cite{RS}&{97.5}\\
\midrule
Triple-CFN&{97.7}\\
\midrule
Triple-CFN + EM&{97.8}\\
\midrule
Triple-CFN+Re-space layer&{98.1}\\
\midrule
Triple-CFN+Re-space layer + EM&\textbf{98.2}\\
\midrule
Meta Triple-CFN&{98.4}\\
\midrule
Meta Triple-CFN+Re-space layer&\textbf{99.3}\\
\bottomrule
\end{tabular}
\end{table}

\begin{table}[h]
\caption{Progressive Pattern Reasoning Accuracies and of Triple-CFN on PGM.}
\label{PGM_Pattern_Results}
\centering
\begin{tabular}{ccccc}
\toprule
&\multicolumn{3}{c}{ Accuracy(\%)}& \\
\cmidrule{2-4}
Model& shape&line&answer \\
\midrule
Meta Triple-CFN&99.5&99.9&98.4\\
\midrule
Meta Triple-CFN+Re-space layer&99.7&99.9&99.3\\
\bottomrule
\end{tabular}
\end{table}

\subsubsection{The generalization task of PGM} Finally, we tested the performance of (Meta) Triple-CFN combined with the Re-space layer on all generalization databases from PGM. We used the specialized dropping process for the Re-space layer. The experimental results, documented in Table \ref{Generalization_PGM_2}, reveal that generalization tasks on reasoning image attributes, such as interpolation, the dropping process of the Re-space layer is both effective and essential. However, this is not the case for generalization tasks on progressive patterns.

\begin{table}[h]

\caption{Generalization Results of Triple-CFN+Re-space layer with Meta Triple-CFN+Re-space layer in PGM.}
\label{Generalization_PGM_2}
\centering
\resizebox{\linewidth}{!}{
\begin{tabular}{cccccccccc}
\toprule
&\multicolumn{3}{c}{Accuracy(\%)}  \\
\cmidrule{2-4}
Dataset&Triple-CFN &Meta Triple-CFN &Meta Triple-CFN \\
&+&+&+Re-space layer\\
&Re-space layer&Re-space layer&(dropping)\\
\midrule
Interpolation & {80.4}&{90.4} &\textbf{94.6}\\
\midrule
Extrapolation & 18.4 & {18.5} &18.6\\
\midrule
Held-out Attribute shape-colour & 12.6 & 13.2 & 12.6 \\
\midrule
Held-out Attribute line-type & 25.2 & 26.8& 21.2 \\
\midrule
Held-out Triples & 22.0 & 23.1& 28.0 \\
\midrule
Held-out Pairs of Triples & 44.5 & \textbf{98.0}& {96.2} \\
\midrule
Held-out Attribute Pairs& 29.2 & \textbf{98.0} & {96.5} \\
\bottomrule
\end{tabular}
}
\end{table}

\subsection{Ablation study}

Previous experiments have already demonstrated the beneficial effects of the dual EM process, Meta process, decorrelated supervision, and Re-space Layer on the CFN framework across datasets such as Bongard-Logo, RAVEN, I-RAVEN, and PGM, thus attesting to their effectiveness. However, as the core focus of this paper, the validity of the CFN framework—specifically, the methodology of separately extracting concepts and features in reasoning problems—warrants further exploration.

To investigate this, we designed the following experiment: we degraded the CFN to its traditional mode, where it only extracts features from images as representations to solve reasoning problems. This required us to adjust the structure of the CFN by removing the $g_\omega(q|x)$ module and replacing the backbone of $g_\eta(z|q,k)$ from a Transformer decoder structure to a Transformer encoder structure to accommodate the absence of the concept $q$. The relevant experimental results are recorded within the entries of CFN$^-$ and Triple-CFN$^-$ in Tables \ref{RAVEN_IRAVEN_Results_}, \ref{PGM_Results_}, and \ref{Bongard_Results_}.
\begin{table}[h]
\caption{Reasoning Accuracies of Triple-CFN$^-$ on RAVEN and I-RAVEN.}
\label{RAVEN_IRAVEN_Results_}
\centering
\resizebox{\linewidth}{!}{
\begin{tabular}{cccccccccc}
\toprule
\toprule
&\multicolumn{8}{c}{Test Accuracy(\%)}& \\
\cmidrule{2-9}
Model&Average&Center&2 $\times$ 2 Grid&3 $\times$ 3 Grid&L-R&U-D&O-IC&O-IG \\
\midrule
Triple-CFN&98.9/{99.1}&100.0/{100.0}&99.7/{99.8}&96.2/{97.5}&99.8/{99.9}&99.8/{99.9}&99.9/99.9&97.0/{97.3} \\
\cmidrule{2-9}
Triple-CFN$^-$&{98.4}/98.5&99.8/{100.0}&{99.7}/{99.6}&{95.4}/96.8&99.4/{100.0}&{99.6}/99.7&{99.9}/99.9&{95.4}/95.5 \\
\bottomrule
\bottomrule
\end{tabular}
}
\end{table}

\begin{table}[hb]
\caption{Reasoning Accuracies of Triple-CFN$^-$ on PGM.}
\label{PGM_Results_}
\centering
\begin{tabular}{ccc}
\toprule
Model&Test Accuracy(\%) \\
\midrule
\midrule
Triple-CFN&{97.7}\\
\midrule
Triple-CFN$^-$& 92.4\\
\bottomrule
\end{tabular}
\end{table}

\begin{table}[hb]
\caption{Reasoning Accuracies of CFN$^-$ and Triple-CFN$^-$ on Bongard-logo.}
\label{Bongard_Results_}
\centering
\begin{tabular}{ccccccc}
\toprule
&\multicolumn{5}{c}{ Accuracy(\%)}& \\
\cmidrule{2-6}
Model&Train& FF&BA&CM&NV \\
\midrule
CFN&92.2&88.5&98.1&77.0&77.3\\
\midrule
CFN$^-$&{91.0}&88.2&98.0&76.2&76.0\\
\midrule
Triple-CFN&93.2&92.0&{98.2}&{78.0}&{78.1}\\
\midrule
Triple-CFN$^-$&{92.5}&{90.5}&{98.1}&{77.5}&{77.0}\\
\bottomrule
\end{tabular}
\end{table}


\section{Conclusion}

Given that abstract reasoning problems pose a significant challenge to the reasoning capabilities of deep learning models, which are widely concerned in the academic community, this paper is dedicated to proposing a series of frameworks that can enhance the capabilities of deep learning models in this regard. This paper makes innovative contributions from multiple dimensions, including the methodology of model structure design, training methods, and the introduction of supervisory signals.
\begin{enumerate}
    \item This paper points out that the difficulty of visual abstract reasoning problems lies not only in the greater challenge of inducting abstract patterns compared to traditional visual discriminative tasks, but also in the conflicts arising from the coexistence of multiple abstract patterns in their low-dimensional representations.
    
    \item Given the existence of such conflicts, this paper emphasizes that the extraction processes of abstract concepts and reasoning features should be explicitly distinguished when designing the architecture of a visual abstract reasoning problem solver. The ablation study conducted in this paper, as well as the superior performance achieved by the CFN (and its advanced version, Triple-CFN) designed based on this methodology, which surpasses that of previous reasoning problem solvers, effectively validates the effectiveness of this methodology.

    \item To more effectively address this conflict, this paper supplements the training process of the CFN framework with a dual EM process. The dual EM process alternately optimizes the parameters of the concept extraction module and the feature extraction module of the CFN, achieving a transformation in the optimization objective of the CFN. This transformation prompts the CFN to actively synthesize a set of concepts on the training data of reasoning problems that are neither conflicting nor alter the problem's solution, and such concept set is the key to mitigating this conflict. The experiments in this paper also demonstrate that the change in optimization objective induced by the dual EM process effectively enhances the performance of the CFN framework.

    \item Coin has two sides. The dual EM process enables CFN to synthesize a set of concepts that overly cater to the training set, which also brings potential Out-Of-Distribution (OOD) risks for these concepts on the test set. To address this, this paper proposes that if the feature extraction process of CFN can retain more reasoning information, such potential risks can be effectively mitigated. Therefore, this paper designs mutual information supervision and decorrelation supervision to assist the feature extraction process of CFN. Experiments on relevant datasets demonstrate the effectiveness of decorrelation supervision.

    \item The dual EM process possesses both effectiveness and limitations. Therefore, this paper attempts to provide explicit supervisory signals directly to the concept extraction process of CFN or Triple-CFN, aiming to achieve the same goals as the dual EM process while avoiding its limitations. This paper notes the metadata that accompanies RPM instances, which can effectively supervise the aforementioned process. Consequently, this paper upgrades Triple-CFN to Meta Triple-CFN by leveraging metadata, achieving remarkable reasoning accuracy and interpretability on the RPM problem.
    
    \item Intuitively and empirically, providing metadata directly to a deep neural network as supplementary supervisory signals to assist in learning RPM problems ought to inherently enhance the network's reasoning accuracy. However, the scenario observed in practice is not the case. Most previous studies have shown that incorporating metadata directly into the training process of an RPM solver can actually decrease its reasoning accuracy \cite{SCL, MRNet, SAVIR-T, RS}. Nevertheless, experimental results related to Meta Triple-CFN demonstrate that it overcomes this phenomenon. 
    This paper attributes this to the structure of Triple-CFN, which separates the concept and feature extraction processes, thereby enabling precise application of metadata supervision to the concept extraction process without interfering with the feature extraction process.

    \item This paper argues that the success of the Meta process in Meta Triple-CFN stems from its construction of a reasonable concept space. Motivated by this, we design the Re-space Layer specifically for constructing a feature space in the feature extraction process. Experimental results demonstrate that the Re-space Layer has the capability to further enhance the reasoning accuracy of Triple-CFN, highlighting the importance of constructing a standardized representation space when building reasoning problem solvers.
    
\end{enumerate}

Overall, this paper presents a series of effective design ideas for abstract reasoning problem solvers. We hope that the successful experiences in designing these solvers can benefit multiple deep learning domains.

\end{document}